\documentclass[lettersize,journal]{IEEEtran}
\usepackage[caption=false,font=normalsize,labelfont=sf,textfont=sf]{subfig}
\usepackage{textcomp}
\usepackage[dvipsnames]{xcolor}

\usepackage[utf8]{inputenc} 
\usepackage[T1]{fontenc}    
\usepackage{url}            
\usepackage{booktabs}       
\usepackage{amsfonts}       
\usepackage{nicefrac}       

\usepackage{graphicx}
\usepackage{amssymb}
\usepackage{pifont}
\usepackage{multirow}
\usepackage{amsmath}
\usepackage{color, colortbl}
\usepackage{caption}

\usepackage{siunitx}

\definecolor{citecolor}{HTML}{0071bc}

\definecolor{Gray}{gray}{0.9}

\newcommand{\KGnote}[1]{{\color{red}{\bf KG: }#1}}

\newcommand{\CAnote}[1]{{\color{brown}{\bf CA: }#1}}

\newcommand{\cmark}{\ding{51}}%
\newcommand{\xmark}{\ding{55}}%

\newcommand{\RNum}[1]{\uppercase\expandafter{\romannumeral #1\relax}}

\renewcommand{\CAnote}[1]{}
\renewcommand{\KGnote}[1]{}

\usepackage{amsmath,amsfonts}
\usepackage{algorithmic}
\usepackage[ruled,linesnumbered]{algorithm2e}
\usepackage{array}
\usepackage[caption=false,font=normalsize,labelfont=sf,textfont=sf]{subfig}
\usepackage{textcomp}
\usepackage{stfloats}
\usepackage{url}
\usepackage{verbatim}
\usepackage{graphicx}
\usepackage{cite}
\usepackage{utfsym}

\def\BibTeX{{\rm B\kern-.05em{\sc i\kern-.025em b}\kern-.08em
    T\kern-.1667em\lower.7ex\hbox{E}\kern-.125emX}}
\usepackage{balance}

\usepackage{times}
\usepackage{fancyhdr,graphicx,amsmath,amssymb}

\usepackage{multirow, bigstrut} 

\usepackage{url}
\usepackage{graphicx}
\usepackage{booktabs} 
\usepackage{xcolor}
\usepackage{hyperref}
\usepackage[capitalize]{cleveref}
\definecolor{deepgreen}{rgb}{0.0, 0.5, 0.0}

\newcommand{\gxlnote}[1]{\textcolor{black}{#1}}

\newcolumntype{P}[1]{>{\centering\arraybackslash}p{#1}}

\begin{document}
\include{pythonlisting}

\title{Redistribute Ensemble Training for Mitigating Memorization in Diffusion Models}


\author{Xiaoliu Guan, Yu Wu, Huayang Huang, Xiao Liu, Jiaxu Miao, Yi Yang
\thanks{X. Guan,  Y. Wu, H. Huang, and  X. Liu are with the School of Computer Science, Wuhan University, China. E-mail: liuxiaoguan, wuyucs, hyhuang, xiaoliu@whu.edu.cn}
\thanks{J. Miao is with the School of Cyber Science and Technology, Sun Yat-sen University, China. E-mail: miaojx@mail.sysu.edu.cn }
\thanks{Y. Yang is with the College of Computer Science and Technology, Zhejiang University, Hangzhou, Zhejiang, China. E-mail: yangyics@zju.edu.cn}
\thanks{(Corresponding author: Yu Wu.)}}

\maketitle

\begin{abstract}
Diffusion models, known for their tremendous ability to generate high-quality samples, have recently raised concerns due to their data memorization behavior, which poses privacy risks. Recent methods for memory mitigation have primarily addressed the issue within the context of the text modality in cross-modal generation tasks, restricting their applicability to specific conditions. In this paper, we propose a novel method for diffusion models from the perspective of visual modality, which is more generic and fundamental for mitigating memorization. 
Directly exposing visual data to the model increases memorization risk, so we design a framework where models learn through proxy model parameters instead. Specially, the training dataset is divided into multiple shards, with each shard training a proxy model, then aggregated to form the final model.
Additionally, practical analysis of training losses illustrates that the losses for easily memorable images tend to be obviously lower. Thus, we skip the samples with abnormally low loss values from the current mini-batch to avoid memorizing.
However, balancing the need to skip memorization-prone samples while maintaining sufficient training data for high-quality image generation presents a key challenge. Thus, we propose IET-AGC+,  which redistributes highly memorizable samples between shards, to mitigate these samples from over-skipping.  
Furthermore, we dynamically augment samples based on their loss values to further reduce memorization.
Extensive experiments and analysis on four datasets show that our method successfully reduces memory capacity while maintaining performance. Moreover, we fine-tune the pre-trained diffusion models, e.g., Stable Diffusion, and decrease the memorization score by 46.7\%, demonstrating the effectiveness of our method. Code is available in \url{https://github.com/liuxiao-guan/IET_AGC}.
\end{abstract}


\section{Introduction}
\IEEEPARstart{R}{ecent} advancements in diffusion models have significantly transformed the landscape of image generation~\cite{croitoru2023diffusion,zhan2023multimodal,zhu2024vision+}. Modern diffusion models, such as Stable Diffusion~\cite{rombach2022high}, Midjourney~\cite{midjourney2022}, and SORA~\cite{sora2024}, can generate realistic images that are hard for humans to distinguish, demonstrating the unparalleled capabilities in producing diverse images. However, recent works~\cite{carlini2023extracting,somepalli2024understanding,wen2023detecting} suggested that diffusion models can memorize images from the training set and reproduce them directly. \gxlnote{This raises privacy concerns, as sensitive information, such as identifiable faces or private documents, may be generated and inadvertently exposed.} To address the critical issue, some works~\cite{zhang2023forget,ni2023degeneration,gandikota2024unified,kumari2023ablating} proposed to make diffusion models ``forget'' specific concepts such as a portrait of a certain celebrity, or the style of a particular artist. However, these works can only blacklist specific content that users want to conceal, but cannot completely cover the privacy-sensitive information that the model might remember, still posing a risk of privacy leakage.

\begin{figure}[tb]
  \centering
  \setlength{\abovecaptionskip}{7pt} 
  \setlength{\belowcaptionskip}{-7pt} 
  \includegraphics[width=1.0\linewidth]{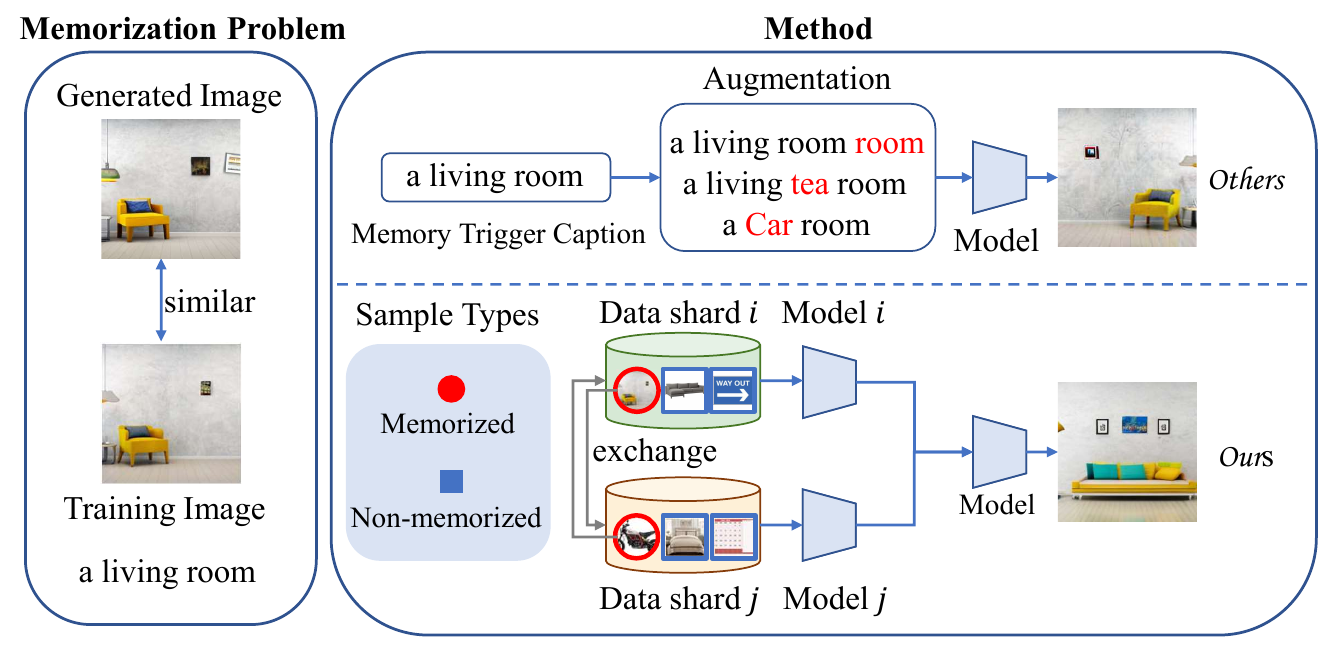}
  \caption{\gxlnote{Prior methods focus solely on the captions associated with the memorized images, such as caption augmentation. In contrast, our approach takes a more generalizable framework by considering aspects from the visual modality.}}
  \label{fig:problem}
\end{figure}
 
Recently, some works~\cite{somepalli2023diffusion,daras2024ambient,somepalli2024understanding,wen2023detecting} have proposed to mitigate diffusion memorization without specific content limitations, thus reducing the risk of diffusion models leaking privacy-sensitive training data. Most of them focused on tackling the training data memorization in text-to-image diffusion models, and proposed data augmentation for captions/sentences to reduce model memorization.  For instance, Somepalli \MakeLowercase{\textit{et al.}}~\cite{somepalli2024understanding} found that the insufficient diversity in captions easily leads to training data generation and thus utilized random caption replacement, random token replacement, and caption word repetition, \MakeLowercase{\textit{etc.}}, to reduce memorization. 
Based on the discovery that memorized prompts tend to exhibit larger magnitudes, which refers to the difference between the text-conditioned and unconditioned noise prediction, Wen \MakeLowercase{\textit{et al.}}~\cite{wen2023detecting} introduced methods for mitigating memorization through filtering high-magnitude sample during training and minimizing magnitudes during inference. Although these works have made significant progress in understanding the memorization issue in diffusion models, \gxlnote{they only focused on easily memorable images related to specific captions in cross-modal generation tasks as shown in ~\cref{fig:problem}. 
However, they do not directly tackle the memorization problem in image generation. While manipulating captions may reduce the likelihood of memorization being triggered in text-to-image models, the model’s inherent ability to memorize images remains. Memorization can still occur under different conditions~\cite{carlini2023extracting,somepalli2024understanding}.
Therefore, we propose a novel framework for diffusion models from the perspective of the visual modality, which not only mitigates memorization more fundamentally but also provides a more generic approach.}

\begin{figure}[tb]
  \centering
  \setlength{\abovecaptionskip}{7pt} 
  \setlength{\belowcaptionskip}{-10pt} 
  \includegraphics[width=1.0\linewidth]{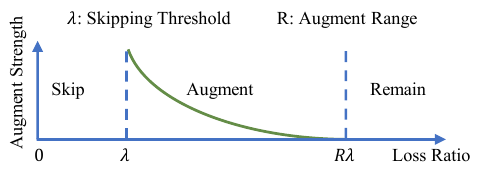}
  \caption{\gxlnote{Threshold-Aware Augmentation (TAA) collaborated with Anti-Gradient Control. We apply three different treatments based on the comparison between the sample's loss ratio and the skipping threshold. }}
  \label{fig:AGC+_s}
\end{figure}

Following these insights,  in our preliminary ECCV 2024 version~\cite{liu2024iterative}, we propose the first module: \textbf{Iterative Ensemble Training (IET)} framework from the perspective of parameter aggregation as shown in ~\cref{fig:problem}. Transmitting data directly to the model increases the likelihood of memorizing easy samples. \gxlnote{However, if the model learns from parameters of other models, rather than directly from the data, it may help to mitigate the direct memorization}. Specifically, we divide the data into multiple data shards and train several proxy models. These models are then aggregated to form the final model. Inspired by federated learning~\cite{mcmahan2017communication}, we iteratively ensemble the proxy models during training, which helps reduce memorization through multiple aggregations and preserves the generation performance. 
Besides, we suspect that images with varying degrees of memorization might exhibit different behaviors during the training process. Therefore, we analyze the training process and find that the loss of easily memorable images tends to be obviously lower than that of less memorable images. Based on this analysis, we propose the second module: \textbf{Anti-Gradient Control (AGC)} to further reduce memorization of training data. In particular, we skip the samples with abnormally small loss values from the current mini-batch to avoid memorizing these samples. During training, as the diffusion model exhibits varying average loss values across different time steps, we maintain a memory bank to track the average loss at each step. Building on this, we skip samples whose loss ratio—defined as the ratio of the sample's loss to the average loss—falls below a predefined skipping threshold as shown in ~\cref{fig:AGC+_s}.

\gxlnote{
However, the AGC strategy might excessively skip highly memorizable samples, leading to a reduction in available training data and potential degradation of image quality.
This drives us to pursue a better approach that strikes a balance between mitigating memorization and maintaining image quality.
Following these insights, in this paper, we introduce IET-AGC+ building on our ECCV2024 framework~\cite{liu2024iterative}.}
\gxlnote{To address the issue of excessively skipping, we propose a \textbf{Memory Samples Redistribute (MSR)} strategy to ensure that these samples are learned but not easily memorized. In the IET framework, each proxy model learns from its shard, where the same data may be interpreted differently.
\emph{In particular, when a sample is frequently memorized in its original shard, it may not have the same memorization tendency in a new shard}. As the saying goes: One man's meat is another man's poison. This inspires us to exchange easily memorized samples from one shard with another to prevent them from being skipped too frequently as shown in ~\cref{fig:problem}.  Therefore, in the training process, we track the number of times each sample is skipped to identify whether it is most easily memorized. During the interaction, each shard allocates its most frequently skipped samples to the next shard in a circular manner.}

\gxlnote{
On the other hand, in AGC, images below the threshold are more likely to be memorized, making their exclusion a reasonable choice. However, memorization varies in degree and cannot be simply addressed with a hard threshold. Samples should be dynamically processed based on their level of memorization risk. To address this, we propose a new strategy called \textbf{Threshold-Aware Augmentation (TAA)} collaborated with Anti-Gradient Control as shown in ~\cref{fig:AGC+_s}. For samples that are not skipped but whose loss values are close to the threshold, we apply augmentation to increase their diversity, thereby reducing memorization. 
A lower loss value indicates a higher risk of memorization, so we use dynamic visual augmentation based on sample distance from the threshold. Samples closer to the threshold receive stronger augmentation.}

Extensive experiments on four datasets highlight the importance of our framework. 
Our method significantly reduces the memorized quantity by \gxlnote{90.1\%, 74.6\%, and 91.2\% }compared with the default training (DDPM~\cite{ho2020denoising}) on CIFAR-10~\cite{krizhevsky2009learning} and CIFAR-100~\cite{krizhevsky2009learning} and AFHQ-DOG~\cite{choi2020stargan}, respectively. Furthermore, when fine-tuning the text-conditional diffusion model, Stable Diffusion~\cite{rombach2022high}, our approach decreases the memorization score by \gxlnote{46.7\%} compared to conventional fine-tuning method~\cite{rombach2022high}. In addition, our method can also be applied to existing inference phase mitigation mechanisms~\cite{somepalli2024understanding,wen2023detecting}, further reducing memorization and improving image quality. These results demonstrate the effectiveness of our method.

\gxlnote{Our main contributions are summarized as follows:
\begin{itemize}
\item {
We introduce a generalized method to mitigate memorization from the perspective of the visual modality, which consists of two main parts: leveraging multiple model ensembles for training and skipping easily memorized samples based on the training loss.
}
\item{
We propose Memory Samples Redistribute (MSR), which redistributes easily memorized samples across shards in the above framework while maintaining a balance between memorization reduction and image quality.
}
\item{
We suggest Threshold-Aware Augmentation (TAA), a strategy that adapts the level of augmentation based on the distance between the sample's loss and the skipping threshold, effectively addressing the risk of overlooking memorized samples.
}
\end{itemize}}

\section{Related Work}

\subsection{Memorization in Generative Models}
Several studies have examined the memorization capabilities of the generative model~\cite{wang2024replication,sun2024create}. 
Generative Adversarial Networks (GANs)~\cite{goodfellow2020generative} have been at the forefront of this research area. 
As Webster \MakeLowercase{\textit{et al.}}~\cite{webster2021person} demonstrated when applied to face datasets, GANs can occasionally replicate.
Prior study~\cite{carlini2021extracting} explored an adversarial attack on language models like GPT-2~\cite{radford2019language}, where individual training examples can be recovered, including personally identifiable information and unique text sequences.

Recent studies have shifted their attention toward diffusion models. 
Somepalli \MakeLowercase{\textit{et al.}}~\cite{somepalli2023diffusion} found that diffusion models accurately recall and replicate training images, especially noted with models like the Stable Diffusion model~\cite{rombach2022high}.
Building upon this discovery, Carlini \MakeLowercase{\textit{et al.}}~\cite{carlini2023extracting} developed a tailored black-box attack for diffusion models. They generated images and implemented a membership inference attack to assess density.
Webster \MakeLowercase{\textit{et al.}}~\cite{webster2023reproducible} demonstrated a more efficient extraction attack with fewer network evaluations, identified "template verbatims," and discussed its persistence in newer systems. 
Recent research has shifted towards exploring the theoretical aspects of memory in diffusion models.
Yoon \MakeLowercase{\textit{et al.}}~\cite{yoon2023diffusion} discovered that generalization and memorization are mutually exclusive occurrences and further demonstrated that the dichotomy between memorization and generalization can be apparent at the class level.
Gu \MakeLowercase{\textit{et al.}}~\cite{gu2023memorization} extensively studied how factors like data dimension, model size, time embedding, and class conditions affect the memory capacity of the diffusion model.

\subsection{Memorization Mitigation} 
The mitigation measures have primarily been concerned with filtering inputs and deduplication. 
For example, Stable Diffusion employed well-trained detectors to identify unsuitable generated content. 
However, these temporary solutions can be easily bypassed~\cite{wen2024hard,rando2022red} and do not effectively prevent or lessen copying behavior on a broad scale. 
Kumari \MakeLowercase{\textit{et al.}}~\cite{kumari2023ablating} designed an algorithm to align the image distribution with a specific style, instance, or text prompt they aim to remove, to the distribution related to a core concept. 
This stopped the model from producing target concepts based on its text condition.
\gxlnote{Hintersdorf \MakeLowercase{\textit{et al.}}~\cite{hintersdorf2024finding} localized memorization of individual data samples down to the level of neurons in DMs’ cross-attention layers.}
However, these approaches are inefficient because they necessitate a list of all concepts to be erased, and have not addressed the key issue of how to reduce the memory capacity of the model.
~\cite{dockhorn2022differentially,ghalebikesabi2023differentially} explored the use of differential privacy (DP)~\cite{dwork2006differential} to train diffusion models or fine-tune ImageNet pre-trained models. However, their focus was on ensuring the privacy of the training of diffusion models, not on the privacy of the images generated by the diffusion models. 
\gxlnote{Chen \MakeLowercase{\textit{et al.}}~\cite{chen2024towards} re-guides generation by measuring the similarity between generated and training images, aiming for memorization-free outputs. However, directly relying on the training set during testing is impractical.}
Daras \MakeLowercase{\textit{et al.}}~\cite{daras2024ambient} introduced a technique for training diffusion models utilizing tainted data. By incorporating additional corruption before applying noise, their methodology prevents the model from overfitting to the training data. But their training requires a considerable amount of time. 
~\cite{somepalli2024understanding,wen2023detecting, ren2024unveiling} also suggested a series of recommendations to mitigate copying such as randomly replacing the caption of an image with a random sequence of words, but most of which are limited to text-to-image models. Our work focuses on the nature of memorization in diffusion models, especially for unconditional ones. 
\vspace{-0.1cm}
\gxlnote{\subsection{Data Augmentation Theory and Practice}
Data augmentation is a widely used technique to improve the generalization of machine learning models, particularly in deep learning ~\cite{wang2021regularizing}. It is commonly employed to increase the diversity of training data by applying transformations in image-based tasks. Common data augmentation techniques include pixel erasing ~\cite{zhong2020random,devries2017improved,chen2020gridmask}, image cropping~\cite{chen2016automatic,ciocca2007self}, mixing images~\cite{hendrycks2019augmix,zhang2017mixup}, geometric transformations~\cite{wang2019perspective,jaderberg2015spatial}, kernel filter~\cite{kang2017patchshuffle}, \MakeLowercase{\textit{etc}}.
The use of data augmentation has been widely explored for vision tasks that require extensive annotation. Azizi \MakeLowercase{\textit{et al.}}~\cite{azizi2023synthetic}showed that augmenting the ImageNet training set~\cite{russakovsky2015imagenet} with samples generated by conditional diffusion models results in a significant boost in classification accuracy. Baranchuk \MakeLowercase{\textit{et al.}}~\cite{baranchuk2021label} investigated how diffusion models can be used to augment data for semantic segmentation, leveraging intermediate activations as rich pixel-level representations, especially when labeled data is scarce. Trabucco \MakeLowercase{\textit{et al.}}~\cite{trabucco2023effective} explored methods to augment individual images with a pre-trained diffusion model, showing significant improvements in few-shot scenarios. Other examples include tasks like human motion understanding~\cite{guo2022learning, izadi2011kinectfusion}, optical flow estimation~\cite{dosovitskiy2015flownet, sun2021autoflow}, and physically realistic simulation environments~\cite{de2022next,dosovitskiy2017carla,gan2020threedworld}, etc. Our study uses data augmentation to flexibly enhance model generalization, thereby mitigating memorization.}

\begin{figure}[t]
  \centering
  \setlength{\abovecaptionskip}{7pt} 
  \setlength{\belowcaptionskip}{-7pt} 
  \includegraphics[height=6.5cm]{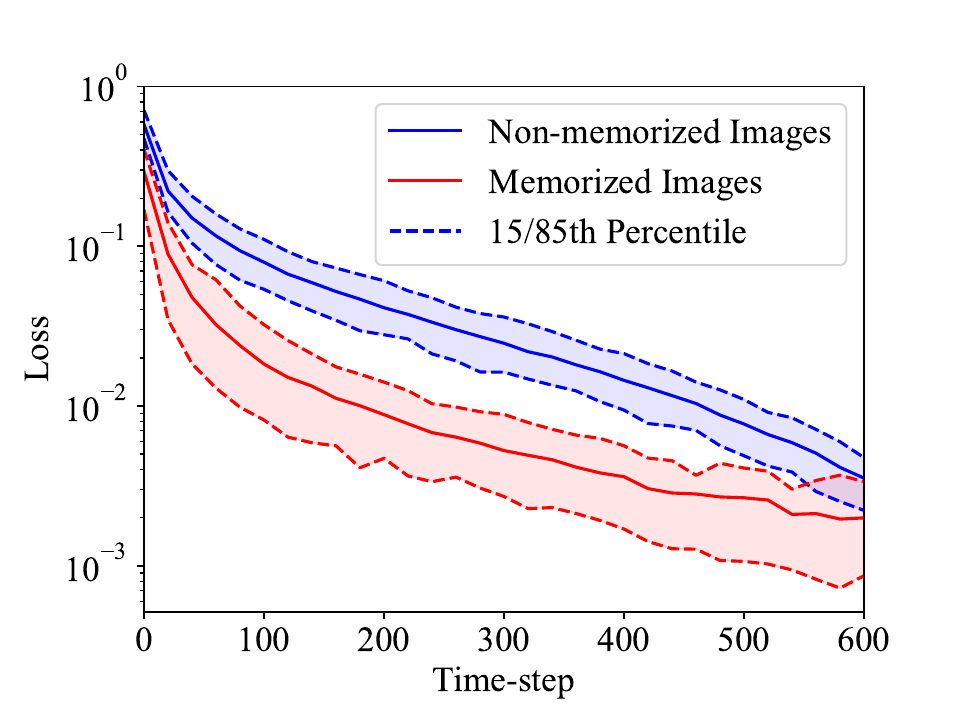}
  \caption{
  Comparison of the training losses between memorized and non-memorized images. 
  }
  \label{fig:LossAnalysis}
\end{figure}
\section{\gxlnote{Exploring Training Loss and Memorization in Diffusion Models}}
To reduce memorization of training data, we delve into the causes of memorization phenomena, specifically analyzing it through the lens of the training loss, \gxlnote{because we suspect that images with varying degrees of memorization might exhibit different behaviors during the training process.}
We begin by establishing the fundamental notation linked with diffusion models.
Diffusion models \cite{ho2020denoising} originate from the non-equilibrium statistical physics \cite{sohl2015deep}.
They are essentially straightforward: they operate as image denoisers.
During the training process, when given a clean image $x$, time-step $t$ is sampled from the interval [$0$, $T$], along with a Gaussian noise vector $\epsilon \sim \mathit{N} (0, I)$,
resulting in a noised image $x_t$:
\begin{equation}\label{eq:noised_data}
    x_t = \sqrt{\alpha_t}x + \sqrt{1-\alpha_t} \epsilon, 
\end{equation}
where the scheduled variance $\alpha_t$ varies between $0$ and $1$, with $\alpha_0 = 1$ and $\alpha_T = 0$. 
The diffusion model then removes the noise to reconstruct the original image $x$ by predicting the noise introduced, achieved through stochastic minimization of the objective function
$\frac{1}{N} \sum_{i} \mathbb{E}_{t,\epsilon} \mathcal{L} (x_i, t, \epsilon; \theta)$, where
\begin{equation}\label{eq:loss}
  \mathcal{L} (x_i, t, \epsilon; \theta) = \| \epsilon - \epsilon_\theta(\sqrt{\alpha_t}x_i + \sqrt{1-\alpha_t}\epsilon, t) \|^2.
\end{equation}

\begin{figure*}[t]
  \centering
  \setlength{\abovecaptionskip}{7pt} 
  \setlength{\belowcaptionskip}{-7pt} 
  \includegraphics[width=1.0\linewidth]{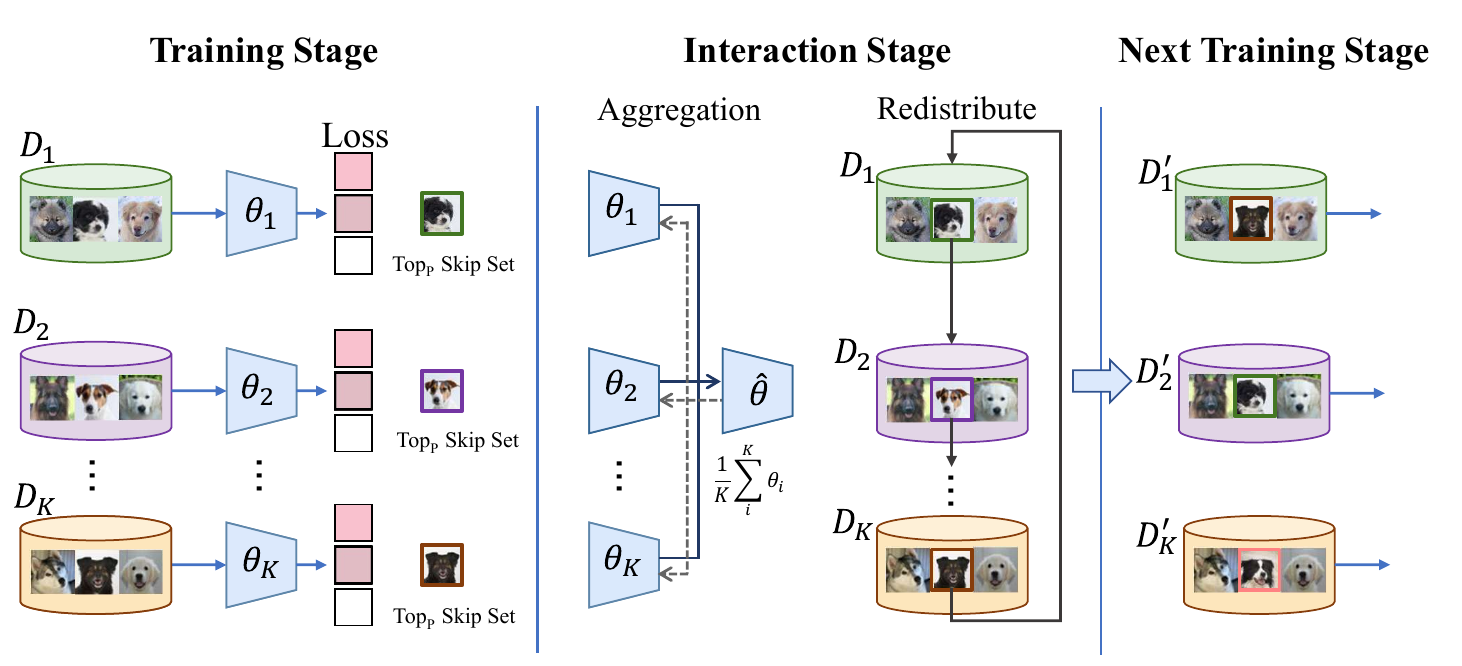}
  \caption{Framework overview of our method. During the training stage, we train multiple proxy models on several data shards. Besides, we selectively skip samples based on their training loss and track how often each sample is skipped in each shard. During the interaction stage, there are two main parts: first, the proxy models are aggregated into a new model, and its weights are distributed as initial weights for the next training phase; second, each shard redistributes its top $P$ skipped samples to the next shard, assigning the last shard to the first. In the next training stage, each shard resumes training with the updated data and model.}
  \label{fig:method}
\end{figure*}
To analyze the correlation between losses and image memorization, \gxlnote{We identify memorized images on CIFAR-10 by generating 65,536 images using a pre-trained model (DDPM)~\cite{ho2020denoising}  and selecting the top 256 training images with the highest similarity to their nearest generated neighbors.} Then we calculate their loss functions at each time step. 
Similarly, we sample 256 non-memorized images from the remaining training data and compute their losses at each time step.
\cref{fig:LossAnalysis} shows the comparisons of the losses.
Memorized images exhibit significantly smaller loss values during this period, indicating that the model tends to reconstruct noise into such images.

\section{Method}
In this section, we present our methodology for mitigating the memorization in diffusion models, without sacrificing excessive image quality.
\subsection{Framework Overview}
\gxlnote{
As shown in ~\cref{fig:method}, our method trains the model by the following two steps iteratively: 
1) training proxy models on each data shard, and 2) conducting two rounds of interaction: proxy model aggregation and shard data redistribution.
Specifically, during the training stage, we divide the dataset into multiple data shards ($D_1, D_2,..., D_K$) and train corresponding proxy diffusion models ($\theta_1, \theta_2,...,\theta_K$). Additionally, we selectively skip certain samples based on their training loss and keep track of the number of times each sample is skipped in each shard. During the interaction stage, the proxy diffusion models  ($\theta_1, \theta_2,...,\theta_K$) from different shards are aggregated into a new model $\hat{\theta}$ through averaging, which serves as the initial model for the next training phase. Meanwhile, each shard identifies and redistributes its top $P$ most easily skipped sample sets to the next shard, updating the data of each shard accordingly. During the next training, each shard resumes training with the updated data shard ($D_1', D_2',..., D_K'$) and model $\hat{\theta}$.
}
\subsection{\gxlnote{Threshold-Aware Control}}
\gxlnote{We first introduce the model updating step.}
In this subsection, we elaborate on how to utilize the aforementioned loss analysis to devise a training strategy to alleviate the occurrence of memorization.
\subsubsection{Anti-Gradient Control}
\textbf{Memory Bank:}
To identify images with exceptionally low loss values that are prone to memorization during training, we need to maintain the average losses for each time step. 
However, computing the average loss at each time step entails substantial computational expenses, as it necessitates evaluating the losses for all images using the model at each time step. Thus, we propose a memory bank to store and update losses during mini-batch training without increasing the time cost. However, 
the losses generally decrease with the training step growing. To address this, when calculating the average loss in the memory bank, we adjust the aggregation process by assigning higher weights to losses that are closer to the current update, rather than simply averaging all losses at the current time step.
Specifically, we initialize an array of length $T$ with zeros, termed the memory bank. 
After calculating the loss for a mini-batch, we update the memory bank using the Exponential Moving Average (EMA) ~\cite{polyak1992acceleration} method based on the loss and the sampled time step, thereby better reflecting the current state of the model:
\begin{equation}\label{eq:ema}
  l_{t} \leftarrow  \eta \cdot l_{t} + (1 - \eta) \cdot \mathcal{L} (x, t, \epsilon; \theta),
\end{equation}
where $\eta$ represents the smoothing factor, and $l_{t}$ represents the averaged loss in the memory bank at time step $t$. 

\textbf{ Loss Ratio-Based Selection:}
In previous observations, if the model exhibits memorization of a certain sample, the loss value of the model on that sample tends to be abnormally small.
Thus, we use the ratio of the training loss of a certain sample to the mean loss in the memory bank at the time step $t$ as a measure to mitigate memorization:
\begin{equation}\label{eq:ratio}
  r(x) = \frac{\mathcal{L} (x, t, \epsilon; \theta)}{l_t}.
\end{equation}
A smaller value of $r(x)$ may indicate a higher likelihood of the image being memorized. 
Then we establish a configurable threshold denoted as $\lambda$. 
If the loss ratio $r(x)$ falls below this threshold $\lambda$, we will skip the image in the mini-batch.
\begin{figure}[tb]
  \centering
  \setlength{\abovecaptionskip}{7pt} 
  \setlength{\belowcaptionskip}{-7pt} 
  \includegraphics[width=1.0\linewidth]{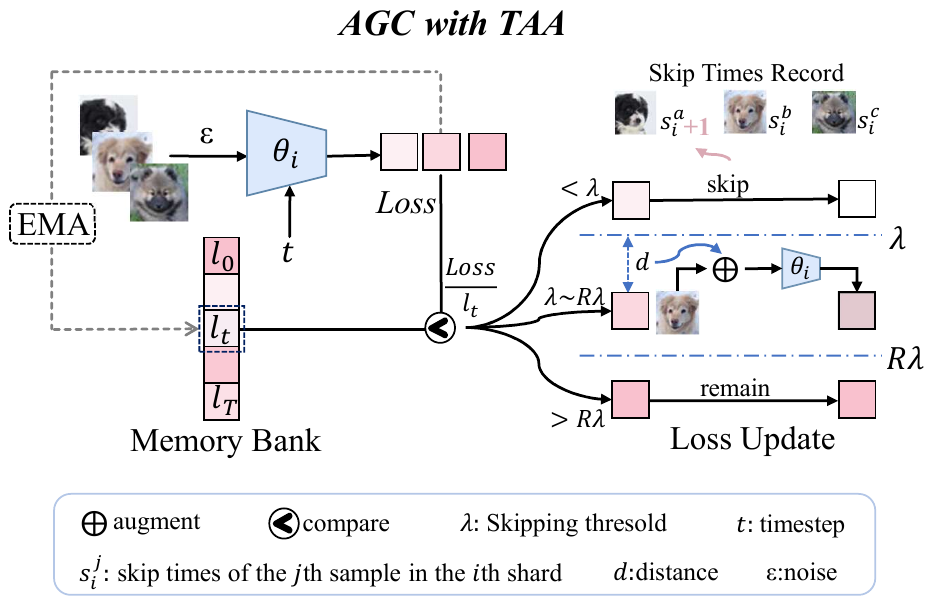}
  \caption{\gxlnote{The proposed model update procedure (AGC with TAA). During training, we dynamically update and maintain a memory bank of losses at each timestep. For each sample's loss ratio $\frac{Loss}{l_t}$, we compare it with 
$\lambda$ and $R\lambda$  to update the loss, considering three cases: for losses less than $\lambda$, we skip the sample and update its skip times; for losses between $\lambda$ and $R\lambda$, we augment the sample and retrain to obtain a new loss; for losses greater than $R\lambda$, we keep the loss unchanged.}}
  \label{fig:AGC+}
\end{figure}

\gxlnote{\subsubsection{Threshold-Aware Augmentation}In AGC, images below the threshold are more likely to be memorized, making their exclusion a reasonable choice. However, memorization varies in degree and samples should be dynamically processed based on their level of memorization risk. Therefore, we design this strategy, dynamically enhancing samples to increase their diversity and thus mitigate memorization.}

\gxlnote{Specifically, for samples not skipped,  if their ratio $r$  does not exceed a specific value, that is, $R$ times the threshold, we apply augmentation to them as follows:
\begin{equation}
\mathcal{L}(\mathbf{Aug}(x, \rho(x), t, \epsilon; \theta) \quad \operatorname{if} \quad \lambda < r(x) <  R\lambda, 
\end{equation}
where $R$ is a multiplier with $R>1$, and $\rho(x)$ represents the relative augmentation strength. 
 For the augmentation function, we choose RandAugment\cite{cubuk2020randaugment} which introduces a vastly simplified search space for data augmentation. 
At the same time, we believe that the lower the sample’s loss value is, the higher its risk of memorization is. Therefore, we apply varying levels of augmentation based on its distance from the threshold—the closer it is, the stronger the augmentation. First, we calculate the relative distance between the loss ratio and the skip threshold:
\begin{equation}
    d(x) = \parallel \frac{ r(x) - \lambda}{\lambda} \parallel.
\end{equation}
Then we choose $e^{-Ax}$ as our negatively correlated function between the distance and the augmentation strength:
\begin{equation}
    \rho(x) =e^{-Ad(x)},
\end{equation}
where $A$ is set  as a constant value of 5.}
\gxlnote{\subsubsection{Threshold-Aware Control}With threshold-aware augmentation, the overall model updating is the following function:
\begin{equation}\label{eq:update_loss_plus}
    \mathcal{L}(x) = 
    \begin{cases} 
    0 & \operatorname{if } r(x) < \lambda \\
    \mathcal{L}(\mathbf{Aug}(x, \rho(x)) & \operatorname{if } \lambda < r(x) < R\lambda \\
    \mathcal{L}(x) & \operatorname{otherwise},
    \end{cases}
\end{equation}
where we re-purpose it by expressing as $\mathcal{L}(x) \propto \mathcal{L}(x, t, \epsilon; \theta)$, omitting $t, \epsilon, \theta$ for simplicity. The overall process is in ~\cref{fig:AGC+}.
}

\subsection{Iterative Ensemble Training}
\gxlnote{In traditional training approaches, directly transmitting the entire training data to the model increases the likelihood of easy samples being memorized. However, if the model learns from parameters of other models, rather than directly from the data, it may help to mitigate memorization. Thus,  we propose a framework that trains multiple proxy diffusion models on different data shards of a dataset. }

\textbf{Training on Different Data Shards.} Unlike the training methods of previous diffusion models, which train a single model on the entire dataset once, in this paper, we divide the dataset into multiple data shards and then train the corresponding proxy diffusion models on each separate part. 
\gxlnote{Specially, we suppose the dataset $D$ contains $N$ samples and $C$ classes.  We divide the dataset into $K$ parts in the IID (Independently and Identically Distributed) setting in which each data shard is randomly assigned a uniform distribution over $C$ classes. 
If the dataset does not contain class information, we divide the dataset into $K$ equal parts. In summary, each data shard contains $\frac{N}{K}$ samples.
}
Then, each shard $i$ trains a separate proxy diffusion model $\theta_{i}$ on its own dataset.

\textbf{Aggregating the Multiple Diffusion Models.} After a period of training, each shard develops a distinct proxy diffusion model. We simply average the weights of all proxy models $\theta_{i}$ to obtain a final model $\hat{\theta}$ as
\begin{equation}\label{eq:average}
    \frac{1}{K}\sum_{i = 1}^{K}\theta_{i} \rightarrow \hat{\theta}.
\end{equation}

Then, we repeat the two stages of training on separate shards of the data and aggregate proxy models, using the obtained final model as the initial model for the first stage.

\textbf{Training Time Analysis.}
As each shard contains only $\frac{1}{K}$ of the total data, the training time for each proxy model is proportionally reduced,
maintaining the overall computational cost \emph{nearly constant} compared to training a single model on the entire dataset. 
The only additional computational cost arises from periodically merging the proxy models, which is minimal and has little impact on overall training efficiency.

\vspace{0.5cm}
\subsection{Memory Samples Redistribute}
\gxlnote{Although AGC effectively mitigates memorization by skipping easily memorized samples, this exclusion may result in reducing the available training data, potentially leading to a decrease in image quality.
To address this issue, we integrated Memory Samples Redistribute (MSR) to ensure that these
samples are learned but not easily memorized.  In the IET framework, each proxy model learns from its shard, where the same data may be interpreted differently. A sample frequently memorized in its original shard may not have the same memorization tendency in a new shard. 
Thus, we allow each shard to redistribute samples that are most easily memorized to the next shard during training, which in practice corresponds to the samples that are most frequently skipped.
}

\gxlnote{Specifically, during the training process, we keep track of the number of times each sample is skipped. We define \( s_i^j \) as skip count for the $j$th sample in the \( i \)th shard's dataset and $s_i = \{s_i^1, s_i^2,..., s_i^\frac{N}{K}\}$ represents the set of skip counts for the $i$th shard.  Then each shard identifies the top $P$ of samples that are most likely to be skipped $s_i^{top}=\{\tilde{s}_i^1, \tilde{s}_i^2,..., \tilde{s}_i^{P*\frac{N}{K}} \}$, where $P$ represents the redistributed proportion of the total samples.  The dataset of most easily memorized samples is defined as:
\begin{equation}
    D_i^{\operatorname{easy}} = \{x^j | s_i^j \in s_i^{top} \}.
\end{equation}
Next, each shard distributes these samples to the next shard in a circular manner, as shown in the following function: 
\begin{equation}
    D_{i + 1} \cup D_i^{\operatorname{easy}}  \rightarrow D_{i + 1}^{\prime}, 
\end{equation}
where $i = 1, 2, \ldots, K.$ }
\gxlnote{
As is shown in ~\cref{fig:method},  the top $P$ most skipped samples from $D_1$ are redistributed to $D_2$, the samples from $D_2$ are assigned to $D_3$, and so on, with the samples from $D_K$ being assigned to $D_1$. In the next training phase, each shard’s dataset is updated accordingly.
}

\section{Experiments}
\subsection{Experimental Setup}
\label{sec:datasets}

\textbf{Datasets.}
We evaluate our method on CIFAR-10~\cite{krizhevsky2009learning}, CIFAR-100~\cite{krizhevsky2009learning}, AFHQ-DOG~\cite{choi2020stargan} for training from scratch, and LAION-10k~\cite{somepalli2024understanding} for fine-tuning text-conditioned model. 
CIFAR-10 and CIFAR-100 consist of 50,000 32x32 color images, divided into 10 and 100 classes respectively. 
AFHQ-DOG is a subset of the AFHQ dataset with approximately 5,000 512x512 dog images, resized to 64x64 for our experiments.
LAION-10k is a subset of LAION~\cite{schuhmann2021laion}, comprising 10,000 image-text pairs with each image having a resolution of 256x256 pixels.
\label{sec:setup}

\textbf{Implementation Details of Training.} 
We conduct experiments on training unconditional diffusion models from scratch using the CIFAR-10, CIFAR-100, and AFHQ-DOG datasets. 
The IET framework divides CIFAR datasets into 10 shards and AFHQ-DOG into 5 shards.
Threshold $\lambda$ is set to 0.5 for CIFAR datasets and 0.7 for AFHQ-DOG. \gxlnote{The augmentation range $R$ is set to 1.7 for CIFAR-10 and CIFAR-100, and 1.2 for AFHQ-DOG.}
To demonstrate the effectiveness of our method in text-conditioned diffusion models, we fine-tune Stable Diffusion~\cite{rombach2022high} on LAION-10k following the setup of Somepalli \MakeLowercase{\textit{et al.}}~\cite{somepalli2024understanding}.
The IET framework divides the LAION-10k dataset into \gxlnote{4} shards, the threshold $\lambda$ is set to 0.8 with \gxlnote{the augmentation range $R$ set to $\infty$.}
For all datasets, the smoothing factor $\eta$ is 0.8, \gxlnote{and the redistribute proportion $P$ is 0.25 }. The augmentation parameter in RandAugment~\cite{cubuk2020randaugment} is set to 5 for CIFAR-100, AFHQ-DOG, and LAION-10k  and 3 for CIFAR-10. 
Further details are in the supplementary material.

\begin{table*}[t]
  \centering
  \caption{Comparisons of unconditional generation on three datasets in terms of memorized quantity denoted as MQ. We also report the FID to evaluate the quality of images produced by the model. Best in bold and second with underline. These notes are the same to other tables following.}
  \begingroup 
  \setlength{\tabcolsep}{5.2pt} 
  \renewcommand{\arraystretch}{1} 
   {\fontsize{7.5}{7.5}\selectfont %
    \begin{tabular}
    {l|l|ccc|c|ccc|c|ccc|c}
     \specialrule{\heavyrulewidth}{0pt}{0pt} 
    \hline
    \multirow{2}[4]{*}{Method} & \multirow{2}[4]{*}{Venue}&\multicolumn{4}{c|}{CIFAR-10} & \multicolumn{4}{c|}{CIFAR-100} & \multicolumn{4}{c}{AFHQ-DOG} \bigstrut\\
\cline{3-14}     &     & MQ$_{0.4}$ & MQ$_{0.5}$ & MQ$_{0.6}$$\downarrow$ & FID$\downarrow$   & MQ$_{0.4}$ & MQ$_{0.5}$ & MQ$_{0.6}$$\downarrow$ & FID$\downarrow$   & MQ$_{0.4}$ & MQ$_{0.5}$ & MQ$_{0.6}$$\downarrow$ & FID$\downarrow$ \bigstrut\\

    \hline
    \hline
    Default (DDPM)~\cite{ho2020denoising}&NeurIPS2020 & 111   & 465   & 2030  & 8.81 & 429   & 1727  & 5620  & 9.29  & 12344 & 19053 & 30795 &23.59 \bigstrut\\
    
     Adding Noise~\cite{ho2020denoising} & NeurIPS2020&197   & 593   & 2091  & 94.61 & 179   & 1037  & 4383  & 86.18  & 1170 0 & 19295 & 27224 & 61.18 \bigstrut\\
   
     Adding DP-SGD~\cite{abadi2016deep} & CCS2016 &148   & 728   & 3200  & 12.55 & -   & -  & -  & -  & - & - & - & - \bigstrut\\
     Ambient Diffusion~\cite{daras2024ambient} & NeurIPS2023& 22   & 138   & 851  & 11.7 & -   & -  & -  & -  & - & - & - & - \bigstrut\\
      
      \hline
       IET-AGC~\cite{liu2024iterative} &ECCV2024& \underline{14}    & \underline{117} & \underline{839}   & \underline{8.34} & \underline{144}  & \underline{760} & \underline{3274}  & \underline{8.51} & \underline{1811}  & \underline{5435} & \underline{15237} & \textbf{22.20} \bigstrut[t]\\
 IET-AGC+ &Ours & \textbf{10}    & \textbf{73} & \textbf{623}   & \textbf{8.33} & \textbf{124}   & \textbf{691} & \textbf{3063}  & \textbf{7.81} & \textbf{1083}  & \textbf{3208} & \textbf{9577} & \underline{24.20} \bigstrut\\
    \hline
     \specialrule{\heavyrulewidth}{0pt}{0pt} 
    \end{tabular}%
    }
    \endgroup
  \label{tab:attack_result}%
  \vspace{-5pt}
\end{table*}%

\textbf{\gxlnote{Evaluation Metrics.}} 
\gxlnote{We evaluate the generations from three perspectives: memorization, generation quality, and text-image alignment.} For memorization, we adopt Carlini's detection rule~\cite{carlini2023extracting} for unconditional generation, considering 
$x$ as memorized if the $\ell_{2}$ distance to its nearest neighbor $\bar{x}$ is significantly lower compared to the $n$ closest neighbors $\mathbb{S}^n_{\bar{x}}$. We modify this rule to:
\begin{equation}
\ell(x,\bar{x};\mathbb{S}^n_{\bar{x}}) = \frac{\ell_{2}(x,\bar{x})}{\mathbb{E}_{y\in{\mathbb{S}^n_{\bar{x}}}}[\ell_{2}(\bar{x},y)]},
  \label{eq:attack}
\end{equation}
where $n = 50$ in our experiment. If the sample's $\ell$-loss value falls below the threshold $\delta_{V}$, it is considered to be memorized:
\begin{equation}
    IsMemo(\delta_{V}, x,\bar{x};\mathbb{S}^n_{\bar{x}}) = \mathbb{I}(\ell_(x,\bar{x};\mathbb{S}^n_{\bar{x}})\leq\delta_{V}).
  \label{eq:attack_threshold}
\end{equation}
The more images below $\delta_{V}$, the stronger the model's memorization.
We generate 65,536 images per model, calculate their $\ell$-loss, and count images below thresholds $\delta_{V}$ of 0.4, 0.5, and 0.6 to quantitatively evaluate the model's memorization, denoted as MQ$_{0.4}$, MQ$_{0.5}$ and MQ$_{0.6}$.
We adopt Somepalli 's evaluation rule~\cite{somepalli2024understanding} for text-conditioned generation, which quantifies memorization using a similarity score derived from the dot product of SSCD features~\cite{pizzi2022self} of $x$ and the nearest neighbor $\bar{x}$:
\begin{equation}
    \zeta = E(\bar{x})^T \cdot E(x),
\end{equation}
where $E(\cdot)$ is the features obtained by SSCD~\cite{pizzi2022self}.  The dataset similarity score (Sim Score) is then defined as the 95th percentile of similarity score distribution for all generated images.
\gxlnote{We use FID~\cite{heusel2017gans} to evaluate the quality of model outputs and Clip Score~\cite{hessel2021clipscore} to measure the generated images’ alignment with the input text prompts.}

\subsection{Experimental Results}
\label{sec:result}
\subsubsection{Training from Scratch}
The experimental results of our method and four competitive methods are shown in Tab.~\ref{tab:attack_result}.
``Default (DDPM)'' denotes the conventional training approach of DDPM~\cite{ho2020denoising}. 
``Adding DP-SGD'' denotes the method of adding Differentially Private Stochastic Gradient Descent~\cite{abadi2016deep}, which involves clipping and adding noise to the model's gradients to protect privacy, albeit at the cost of some image quality.
``Adding Noise'' denotes a method of directly adding Gaussian noise to the images during training, with a mean of 0 and a variance of 0.1.
``Ambient Diffusion''~\cite{daras2024ambient} protected privacy by training generative models on highly corrupted samples, preventing the model from directly observing clean training data. ``IET-AGC'' is our preliminary version ~\cite{liu2024iterative}.

Results in Tab.~\ref{tab:attack_result} show that adding noise or gradients to the training images reduces the quality of the generated images. 
However, it still does not resolve the issue of training image memorization.
Despite Ambient Diffusion also reducing memorization, it leads to a significant increase in FID (from 8.81 to 11.7), indicating a notable degradation of image quality.
Compared with the default training approach, our method maintains or even slightly improves the generative quality by reducing the FID score.
At the same time, our method significantly reduces the diffusion model's memorization of the training data. As shown in Tab.~\ref{tab:attack_result}, for the MQ$_{0.4}$ score, the number of memorized images reduces by \gxlnote{90.1\%, 74.6\%, and 91.2\%} compared with the default training on CIFAR-10, CIFAR-100, and AFHQ-DOG, respectively, illustrating the effectiveness of our method.

\subsubsection{Fine-tuning Pre-trained Diffusion Models}
\label{sec:finetune}
Training a diffusion model from scratch requires a significant amount of computational resources and time. Thus, fine-tuning a pre-trained diffusion model with limited epochs to reduce memorization is necessary.
To further demonstrate the effectiveness and applicability of our method, we finetune text-conditional Stable Diffusion.

For baselines, we compare the methods from ``Default (SD)'', Somepalli \MakeLowercase{\textit{et al.}}~\cite{somepalli2024understanding}, and Wen \MakeLowercase{\textit{et al.}}~\cite{wen2023detecting}.
``Default (SD)'' denotes the conventional fine-tuning approach of SD~\cite{rombach2022high}.
The results are presented in \cref{tab:finetnue_SD}.
Somepalli \MakeLowercase{\textit{et al.}}~\cite{somepalli2024understanding} protected privacy by randomizing conditional information (e.g., RT, CWR, GNI, MC, RC, and CWR in \cref{tab:finetnue_SD}) during training and inference, thereby reducing the likelihood of the model replicating specific training data. 
\gxlnote{Wen \MakeLowercase{\textit{et al.}}~\cite{wen2023detecting} also mitigated memorization in two stages: excluding samples exceeding a certain threshold during training and adjusting prompt embeddings during inference. 
The method proposed by Somepalli \MakeLowercase{\textit{et al.}}~\cite{somepalli2024understanding} has limited effectiveness in mitigating memorization, both during the training phase and the inference phase.
On the other hand, the approaches designed by Wen \MakeLowercase{\textit{et al.}}~\cite{wen2023detecting} achieve high performance in Sim Score but excessively excluding samples limits improvements in text alignment and image quality.
However, our method effectively balances memorization and generation quality, achieving a Sim Score of 0.34, which represents a 46.7\% reduction compared to the default method, while maintaining the highest Clip Score of 31.27 and competitive FID of 16.3. }
\begin{table}[t]
  \centering
  \caption{Fine-tuning results of Stable Diffusion model on LAION-10k. ``Phase'' refers to the phase for mitigating memorization, encompassing both the inference phase and the training phase.}
  \begingroup 
  \setlength{\tabcolsep}{3.5pt} 
  \renewcommand{\arraystretch}{1} 
    \begin{tabular}{l|ll|l|ccc}
    \specialrule{\heavyrulewidth}{0pt}{0pt} 
    \hline
    Phase & \multicolumn{2}{l|}{Method} &Venue& Sim Score↓ & Clip Score↑ & FID↓ \bigstrut\\
    \hline
    \hline
          & \multicolumn{2}{l|}{Default (SD)~\cite{rombach2022high}} &CVPR2022& 0.638  & 30.52 & 18.7  \bigstrut\\
    \hline
    \multirow{4}[2]{*}{Infer.} & \multicolumn{2}{l|}{RT~\cite{somepalli2024understanding}}&NeurIPS2023 & 0.524  & 29.54 & 18.7  \bigstrut[t]\\
          & \multicolumn{2}{l|}{CWR~\cite{somepalli2024understanding}}&NeurIPS2023 & 0.576  & 30.13 & 18.1  \\
          & \multicolumn{2}{l|}{GNI~\cite{somepalli2024understanding}}&NeurIPS2023 & 0.615  & 30.32 & 18.9  \\
          & \multicolumn{2}{l|}{Wen \MakeLowercase{\textit{et al.}}~\cite{wen2023detecting}}&ICLR2024 & 0.352  & 28.56 & 25.7  \bigstrut[b]\\
    \hline
    \multirow{6}[4]{*}{Train} & \multicolumn{2}{l|}{MC~\cite{somepalli2024understanding}} &NeurIPS2023& 0.420  & 30.27 & 16.6  \bigstrut[t]\\
           & \multicolumn{2}{l|}{RC~\cite{somepalli2024understanding}}&NeurIPS2023 & 0.565  & 30.64 & \textbf{16.0} \\
          & \multicolumn{2}{l|}{CWR~\cite{somepalli2024understanding}}&NeurIPS2023 & 0.614  & 30.79 & 16.7  \\
          & \multicolumn{2}{l|}{Wen \MakeLowercase{\textit{et al.}}~\cite{wen2023detecting}}&ICLR2024 & \textbf{0.320} & \underline{30.86}  & 17.5  \bigstrut[b]\\
\cline{2-7}          & \multicolumn{2}{l|}{IET-AGC~\cite{liu2024iterative}}&ECCV2024 & 0.393  & 31.25  & 16.9 \bigstrut[t]\\
          & \multicolumn{2}{l|}{IET-AGC+} &Ours& \underline{0.340}  & \textbf{31.27} & \underline{16.3}  \bigstrut[b]\\
    \hline
    \specialrule{\heavyrulewidth}{0pt}{0pt} 
    \end{tabular}%
    \endgroup
  \label{tab:finetnue_SD}%
  \vspace{-12pt}
\end{table}%

\begin{figure}[t]
  \centering
  \setlength{\abovecaptionskip}{7pt} 
  \setlength{\belowcaptionskip}{-2pt} 
  \includegraphics[width=1.0\linewidth]{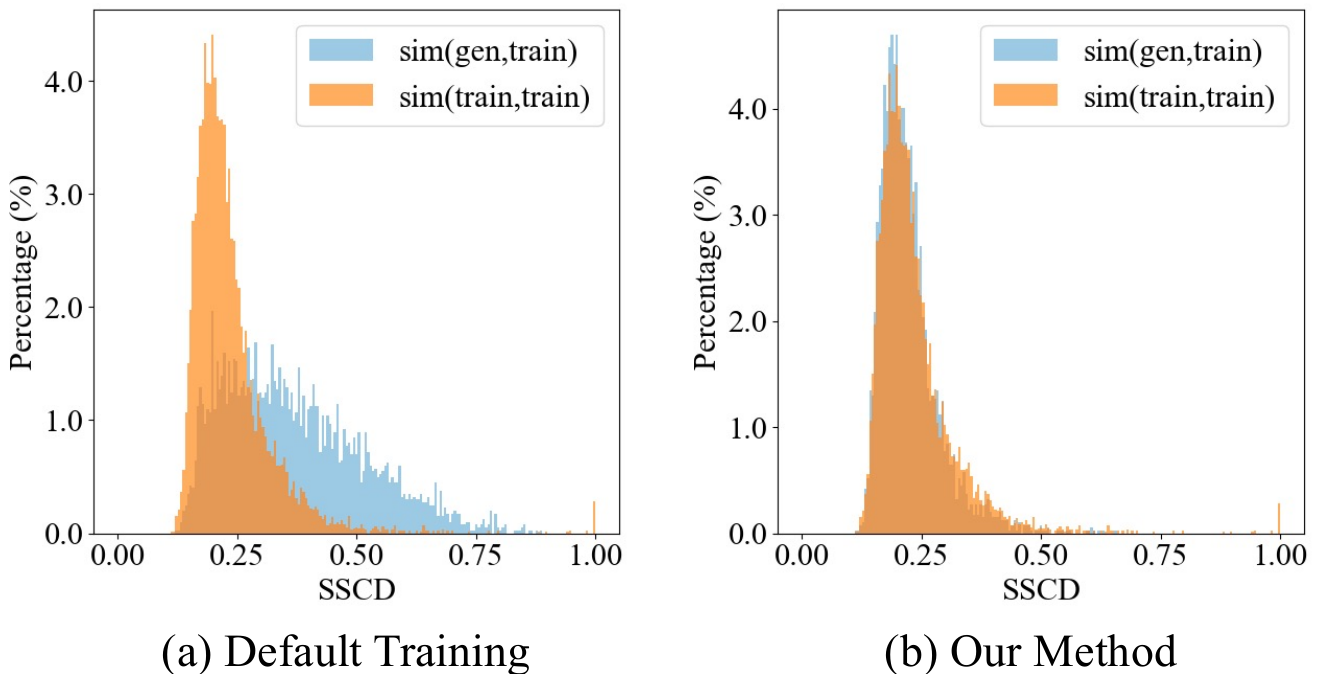}
  \caption{\gxlnote{Comparison of the Sim Score histograms between the generated images and the training images for both the default method and our approach. The label sim(train, train) refers to the Sim Score between images in the training set and all other training images (excluding the image itself).}}
  \label{fig:SD_histogram}
\end{figure}

\gxlnote{Additionally, we also present similarity score distribution plots of all generated images in ~\cref{fig:SD_histogram}. Compared to the default method, our approach results in overall lower similarity scores, with the majority of similarity scores of the data concentrated in the 0.2$\sim$0.3 range, showing closer similarity to the training set itself. This further demonstrates that our method significantly reduces the model's memorization ability.}

\begin{figure}[t]
    \centering
    \setlength{\abovecaptionskip}{12pt} 
  \setlength{\belowcaptionskip}{2pt} 

    \begin{minipage}{0.2\linewidth}
        \centering
        Training Image
    \end{minipage}%
    \hspace{1em} 
    \begin{minipage}{0.7\linewidth}
        \centering
        \includegraphics[width=\linewidth]{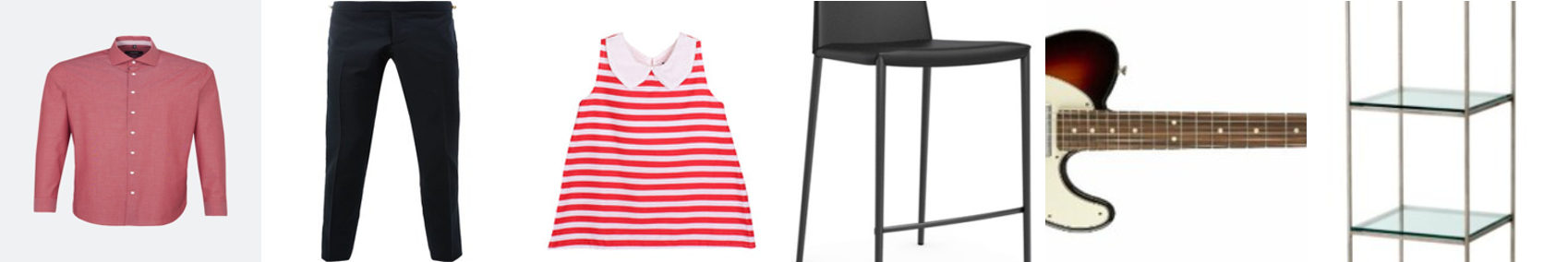} 
    \end{minipage}

    \vspace{1em}

    \begin{minipage}{0.2\linewidth}
        \centering
    Default (SD)~\cite{rombach2022high} 
    \end{minipage}%
    \hspace{1em} 
    \begin{minipage}{0.7\linewidth}
        \centering
        \includegraphics[width=\linewidth]{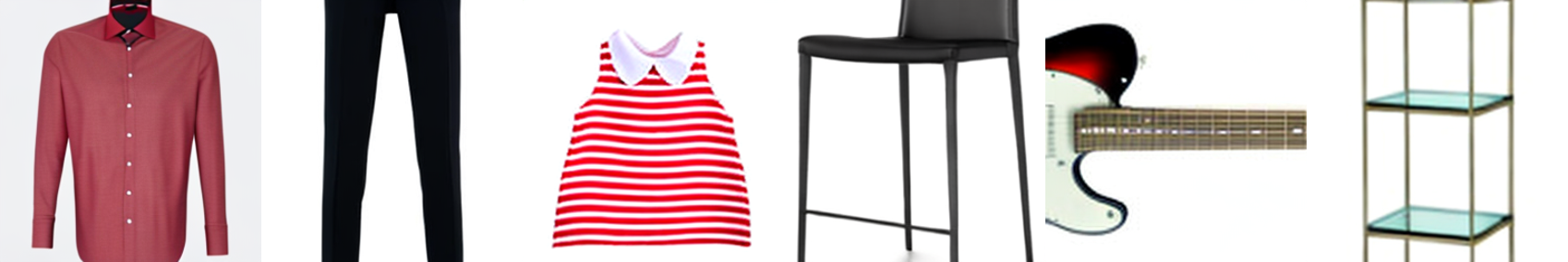} 
    \end{minipage}
    
    \vspace{1em}
    \begin{minipage}{0.2\linewidth}
        \centering
        CWR~\cite{somepalli2024understanding}
    \end{minipage}%
    \hspace{1em} 
    \begin{minipage}{0.7\linewidth}
        \centering
        \includegraphics[width=\linewidth]{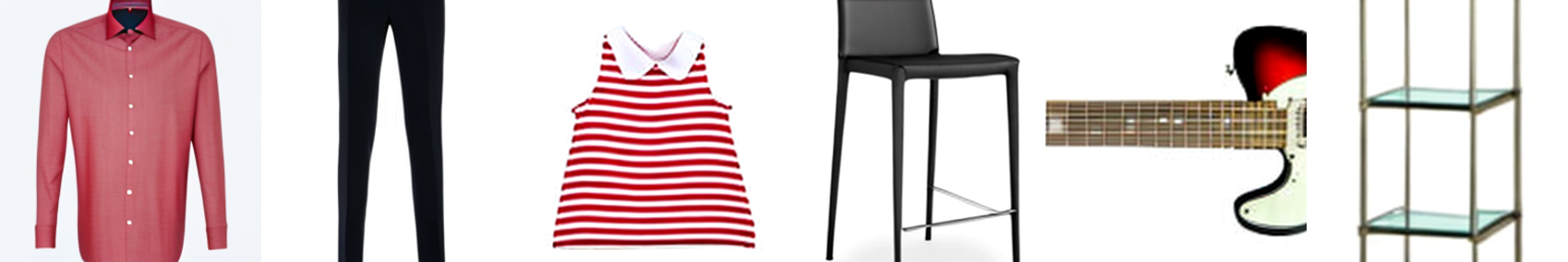} 
    \end{minipage}
    
    \vspace{1em}
    \begin{minipage}{0.2\linewidth}
        \centering
        RT~\cite{somepalli2024understanding}
    \end{minipage}%
    \hspace{1em} 
    \begin{minipage}{0.7\linewidth}
        \centering
        \includegraphics[width=\linewidth]{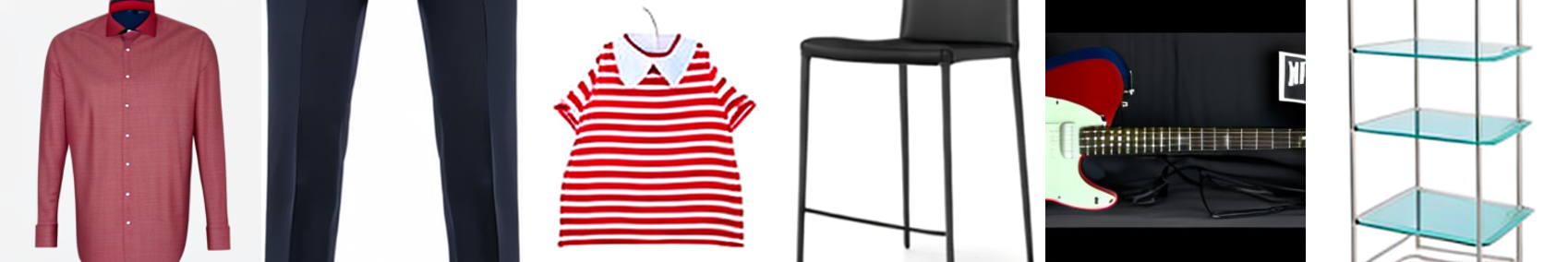} 
    \end{minipage}

    \vspace{1em}
    \begin{minipage}{0.2\linewidth}
        \centering
        Wen \MakeLowercase{\textit{et al.}}~\cite{wen2023detecting}
    \end{minipage}%
    \hspace{1em} 
    \begin{minipage}{0.7\linewidth}
        \centering
        \includegraphics[width=\linewidth]{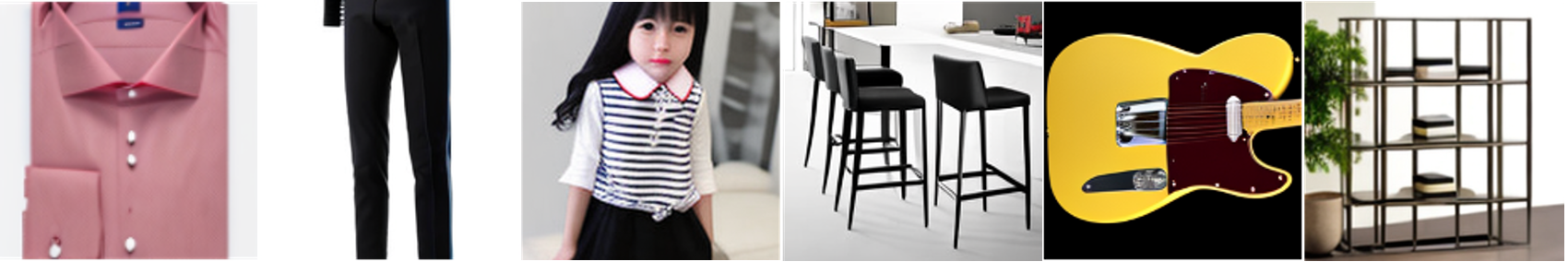} 
    \end{minipage}

    \vspace{1em}
    \begin{minipage}{0.2\linewidth}
        \centering
        Ours
    \end{minipage}%
    \hspace{1em} 
    \begin{minipage}{0.7\linewidth}
        \centering
        \includegraphics[width=\linewidth]{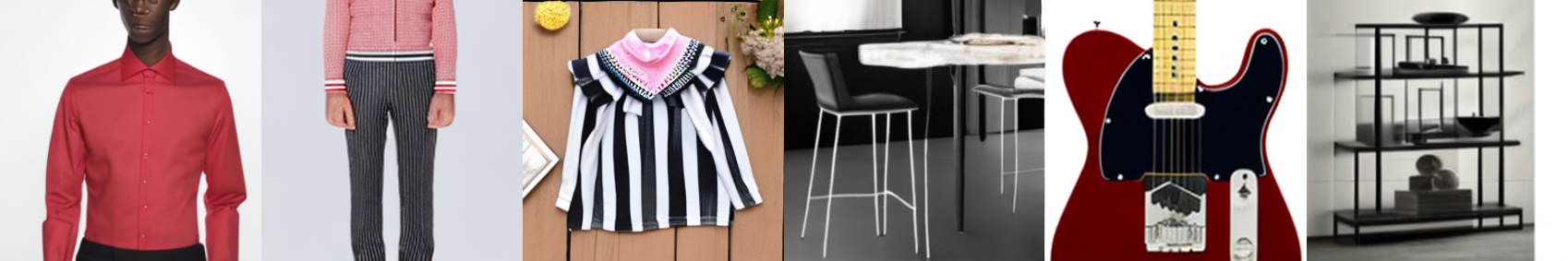} 
    \end{minipage}
    \caption{The visualizations of the generated images from our method and the baseline methods. Each column presents images generated by different methods using the same caption and random seed, alongside the corresponding training set images for that caption.}
    \label{fig:Visual}
    \vspace{-1em}
\end{figure}

\textbf{Visualization.} To provide a more intuitive confirmation of our training method, with the conditions of the same captions, we visualize the images generated from our
method and the baseline methods  \cref{fig:Visual}.
\gxlnote{When the method without mitigation is applied, the generated images exhibit high similarity to the training images. While memorization mitigation methods show some differences from the training images, the effect is not as pronounced or the quality of the generated images slightly decreases. In contrast, the images generated by our method are more diverse in content, and their quality remains high without significant degradation.}

\subsection{Analysis of Skipping}
\label{sec:analysis}
In this section, we conduct comparative experiments on the AFHQ-DOG dataset to delve into which types of images are prone to be skipped, as well as the relationship between memorizable images and the images that are skipped.

\subsubsection{Images Most Easily Skipped}
\begin{figure*}[tb]
  \setlength{\abovecaptionskip}{0pt} 
  \setlength{\belowcaptionskip}{-7pt} 
  \centering
  \begin{minipage}{0.43\linewidth}
    \includegraphics[width=1.\linewidth]{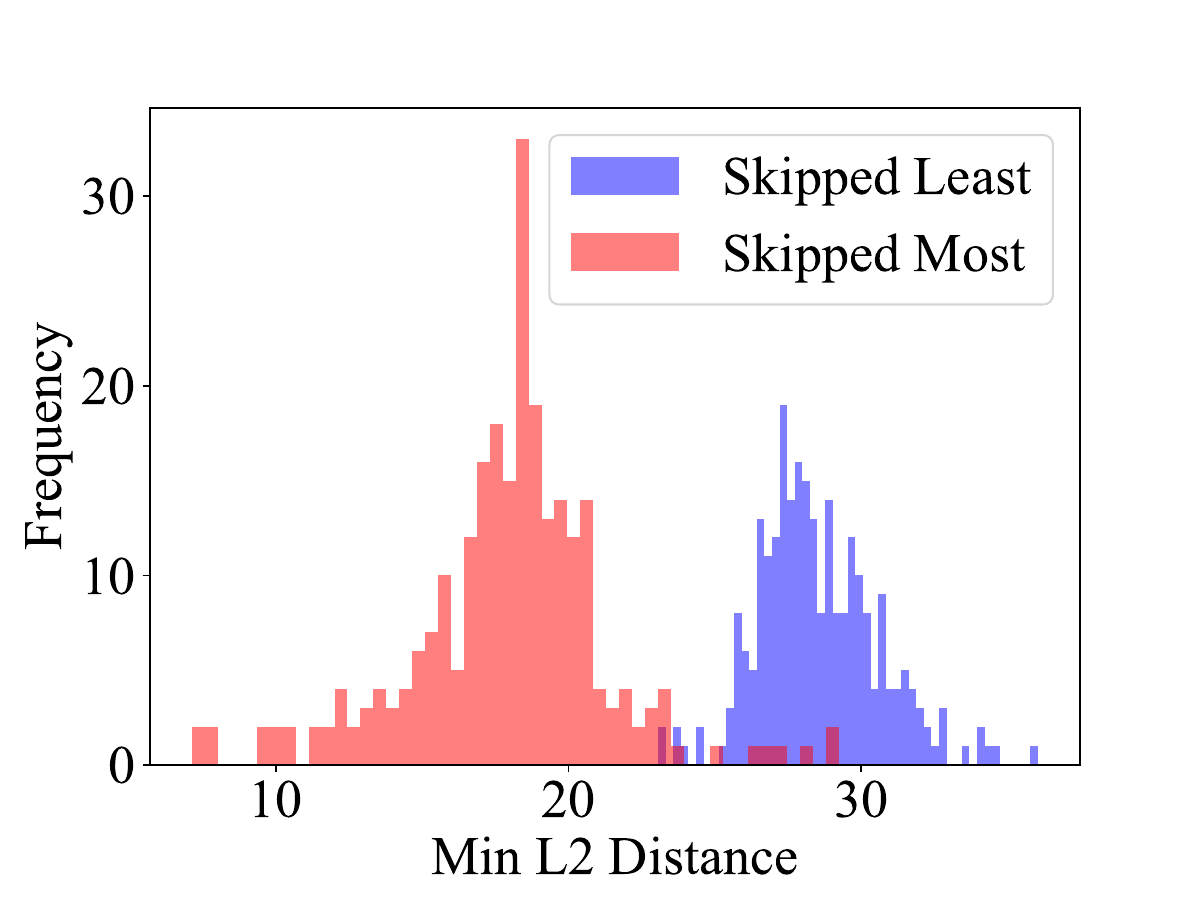}
    \label{fig: Ana_l2_most_least}
  \end{minipage}
  \begin{minipage}{0.43\linewidth}
    \includegraphics[width=1.\linewidth]{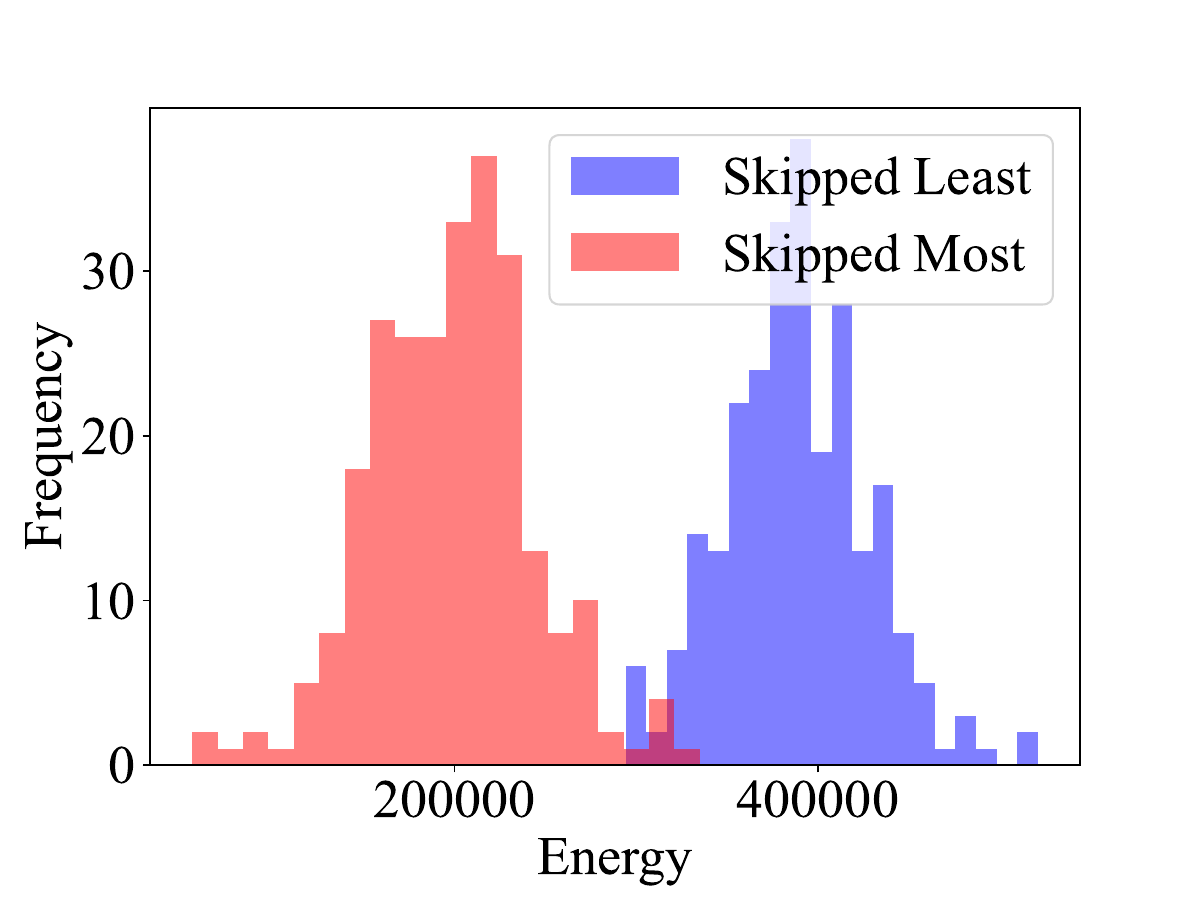}
    \label{fig: Ana_spec}
  \end{minipage}
  \caption{The data distribution analysis of images skipped most and images skipped least. The left subplot shows the distribution of distances to the most similar images in the dataset. The right subplot displays energy distribution. The greater the energy, the more complex the image.}
  \label{fig:Ana_hist}
\end{figure*}

We believe the images are more easily skipped for two main reasons.
Firstly, data aggregation: we compute the $\ell_{2}$ distance between these easily skipped images and all other images in the dataset, as well as between those not easily skipped images and all other images in the dataset.
The left subplot in \cref{fig:Ana_hist} indicates that the distribution of the skipped images is more clustered.
Consistent with the findings of Carlini \MakeLowercase{\textit{et al.}}~\cite{carlini2023extracting}, which suggested that removing duplicate training images effectively reduces memorization capacity, skipping these clustered images can also reduce memorization capacity.
Secondly, data simplicity: we performed Fourier transforms~\cite{sneddon1995fourier} on these easily skipped and not easily skipped images to obtain their energy distributions. This process helps decompose the image into different frequency bands, where low frequencies correspond to broad, smooth structures, and high frequencies capture fine details or noise. By examining the frequency spectrum, we quantified the energy distribution, which reflects the amount of information or complexity present in the image. As is shown in the right subplot of \cref{fig:Ana_hist},
the easily skipped images have less energy, indicating that they lack finer details.
We believe both factors contribute to the model’s tendency to memorize these images, making their skipping effective in reducing memorization capacity.

\subsubsection{Frequency of Skipped Images}
\begin{figure}[tb]
  \setlength{\abovecaptionskip}{-0pt} 
  \setlength{\belowcaptionskip}{-7pt} 
  \centering
  \includegraphics[height=6.5cm]{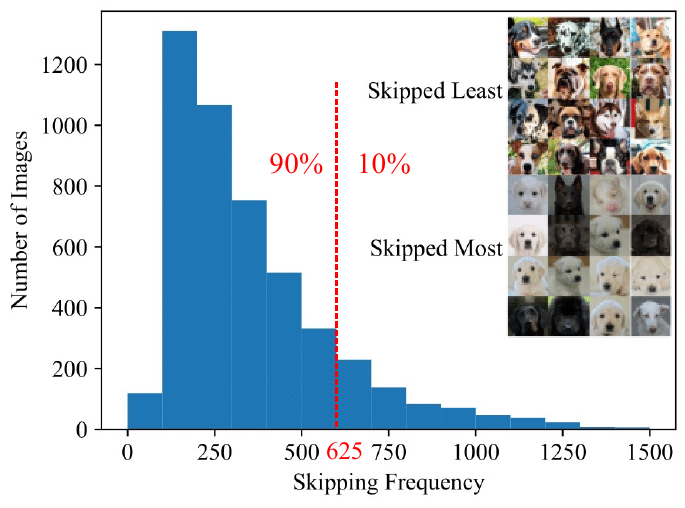}
  \caption{
    Distribution of skipped image counts. There are about 90\% of the images are skipped fewer than 625 times. 
  }
  \label{fig:value_counts}
\end{figure}

Throughout the training process, we record the identifiers of skipped images. 
As shown in \cref{fig:value_counts}, our method does not entail skipping all images. 
In our approach, about 90\% of the images are skipped fewer than 625 times (across a total of 2,278 training epochs), indicating that our method can effectively differentiate between different images. 
This suggests that we are not simply reducing memorization by constraining the model's learning. 
On the other hand, while our method requires skipping images with exceptionally low loss values, all images still contribute to the model's training.

\subsection{Ablation Study}
\label{sec:ablation}
\subsubsection{Performance Comparisons of Each Component}
To further understand the effectiveness of our approach, we conduct ablation experiments to investigate the individual impacts of different components for training from scratch and fine-tuning Stable Diffusion on LAION-10k. 

\gxlnote{\textbf{Effectiveness of AGC:}~\cref{tab:ablation_cifar10} and ~\cref{tab:ablation_SD} show that Anti-Gradient Control (AGC) effectively mitigates model memorization by excluding easily memorized samples, in both training from scratch and fine-tuning scenarios. When training from scratch,  AGC reduces MQ$_{0.5}$ (465 to 154) by approximately 67\% compared to the conventional method. However, excessive exclusion of samples can reduce the number of training images, which in turn impacts the quality of the generated images, as evidenced by the improvement in FID shown in the ~\cref{tab:ablation_cifar10}.}

\gxlnote{\textbf{Effectiveness of IET:} For Iterative Ensemble Training (IET), it is evident that the way the model learns from the parameters of the proxy model can not only reduce memorization but also significantly improve the quality of images in ~\cref{tab:ablation_cifar10} and ~\cref{tab:ablation_SD}. When training from scratch, although it is not as effective as AGC in reducing memorization, it greatly reduces FID. Compared with AGC, FID has decreased by 26.6\%, greatly improving the quality of images. When fine-tuning, IET has the same effect. Compared with the default method, the Clip Score increases from 30.52 to 31.27. }
\begin{table}[t]
  \centering
  \caption{Performance comparisons of each component for \emph{training from scratch}. }
  {\fontsize{7.5}{7.5}\selectfont
    \begin{tabular}{cccc|ccc|c}
    \specialrule{\heavyrulewidth}{0pt}{0pt} 
    \hline
    \multicolumn{4}{c|}{Method}   & \multicolumn{4}{c}{CIFAR-10} \bigstrut\\
    \hline
    AGC   & IET   & TAA   & MSR   &  MQ$_{0.4}$ & MQ$_{0.5}$ & MQ$_{0.6}$$\downarrow$ & FID$\downarrow$ \bigstrut\\
    \hline
    \hline
    \xmark     & \xmark     & \xmark     & \xmark     & 111   & 465   & 2030  & 8.81 \bigstrut[t]\\
    \cmark     & \xmark     & \xmark     & \xmark     & 26    & 154   & 976   & 11.36 \\
    \cmark     & \cmark     & \xmark     & \xmark     & 14   & 117   & 839   & \underline{8.34} \\
    \cmark     & \cmark     & \cmark     & \xmark     & \textbf{8} & \underline{81} & \underline{678} & 9.20 \\
    \cmark     & \cmark     & \cmark     & \cmark     & \underline{10}     & \textbf{73}    & \textbf{623}   & \textbf{8.33} \bigstrut[b]\\
    \hline
    \specialrule{\heavyrulewidth}{0pt}{0pt} 
    \end{tabular}%
    }
  \label{tab:ablation_cifar10}%
  \vspace{-7pt}
\end{table}%

\gxlnote{\textbf{Effectiveness of TAA:} Considering that memorization varies in degree and cannot be simply addressed with a hard threshold, we propose Threshold-Aware Augmentation (TAA). The fourth rows of ~\cref{tab:ablation_cifar10}  and ~\cref{tab:ablation_SD} show that augmenting samples above the skipping threshold effectively mitigates the issue of overlooked memorized samples in the AGC strategy and further reduces memorization. In training from scratch, compared with our conference version method, MQ$_{0.5}$ is decreased from 117 to 81 by 30.8\%. 
At the same time, by applying varying levels of augmentation based on the sample's loss, TAA does not reduce image quality to a noticeable extent. As for fine-tuning, compared with IET-AGC,  Clip Score is increased from 31.25 to 31.18.}

\gxlnote{\textbf{Effectiveness of MSR:} For Memory Samples Redistribute (MSR), we can see that in ~\cref{tab:ablation_cifar10} and ~\cref{tab:ablation_SD}, re-by engaging excessively skipped samples for learning, the MSR method significantly enhancing image quality. When training from scratch, compared to the previous row, FID is decreased from 9.20 to 8.33.  In addition,  the model's memorization capacity is not significantly affected. This suggests that once the easily memorized samples are exchanged across shards, they no longer retain their high memorization potential. For instance, when training from scratch, compared to the previous row, MQ$_{0.6}$ is decreased from 678 to 623.}

\begin{table}[t]
  \centering
  \caption{Performance comparisons of each component for \emph{fine-tuning} Stable Diffusion on LAION-10k.}
    \begin{tabular}{cccc|ccc}
    \specialrule{\heavyrulewidth}{0pt}{0pt} 
    \hline
    \multicolumn{4}{c|}{Method}   & \multirow{2}[4]{*}{Sim Score↓} & \multirow{2}[4]{*}{Clip Score↑} & \multirow{2}[4]{*}{FID↓} \bigstrut\\
\cline{1-4}    AGC   & IET   & TAA   & \multicolumn{1}{c|}{MSR} &       &       &  \bigstrut\\
    \hline
    \hline
     \xmark     & \xmark     & \xmark     & \xmark    & 0.638  & 30.52 & 18.7  \bigstrut[t]\\
    \cmark     & \xmark     & \xmark     & \xmark     & 0.533  & 30.57  & 18.5 \\
    \cmark     & \cmark     & \xmark     & \xmark    & 0.393  & 31.25  & 16.9 \\
   \cmark     & \cmark     & \cmark     & \xmark      & 0.350  & 31.18 & 16.7 \\
    \cmark     & \cmark     & \cmark     & \cmark      & \textbf{0.340} & \textbf{31.27} & \textbf{16.3} \bigstrut[b]\\
    \hline
    \specialrule{\heavyrulewidth}{0pt}{0pt} 
    \end{tabular}%
  \label{tab:ablation_SD}%
\end{table}%

\subsubsection{Different Samples Redistribute Methods}
MSR is designed to address the issue of excessive skipping and then to improve image quality.
To further validate the effectiveness of MSR, we experiment with random samples redistribute, where samples exchanged between shards are selected randomly. The results, shown in ~\cref{tab:ablation_MSR}, indicate that there is little difference in memory mitigation between the two redistribution methods. In contrast, memory samples redistribute achieves superior image quality, demonstrating that frequently skipped samples are essential for improving image generation quality. MSR facilitates their relearning, effectively enhancing overall performance. However, random samples redistribute lacks specificity in addressing such samples, resulting in no significant improvement in image quality.
\begin{table}[t]
  \centering
  \caption{The result about different samples redistribute strategies in our method. The suffix ``Random'' represents random samples redistribute, where samples exchanged between shards are selected randomly. The suffix ``Memory'' represents memory samples redistribute, \MakeLowercase{\textit{i.e.}}, MSR.}
    \begin{tabular}{ll|ccc}
    \specialrule{\heavyrulewidth}{0pt}{0pt} 
    \hline
    \multicolumn{2}{l|}{Method} & Sim Score↓ & Clip Score↑ & FID↓ \bigstrut\\
    \hline
    \hline
    \multicolumn{2}{l|}{IET-AGC+$_\mathnormal{Random}$} & \textbf{0.340} & 31.10  & 16.60  \bigstrut[t]\\
    \multicolumn{2}{l|}{IET-AGC+$_\mathnormal{Memory}$} & \textbf{0.340} & \textbf{31.27} & \textbf{16.30} \\
    \hline
    \specialrule{\heavyrulewidth}{0pt}{0pt} 
    \end{tabular}%
  \label{tab:ablation_MSR}%
\end{table}%

\subsubsection{Composition of our methods with existing works}
\gxlnote{Our approach is orthogonal to existing state-of-the-art mitigation works. To demonstrate the applicability of our method, we apply our method to  Somepalli \MakeLowercase{\textit{et al.}}~\cite{somepalli2024understanding} and Wen \MakeLowercase{\textit{et al.}}~\cite{wen2023detecting} 's inference phase mitigation mechanisms. The results are shown in ~\cref{tab:addinference}. Our method can be applied to their approach to further enhance performance. Not only does it reduce memorization, but also it improves image quality and text alignment. For instance, ``GNI+Ours'' shows a 45.2\% decrease in Sim Score compared to ``GNI'', a 0.54 (30.32 to 30.86) increase in Clip Score, and a 1.9 (18.9 to 17.0) reduction in FID. }


\begin{table}[t]
  \centering
  \caption{Composition of our methods with existing works. Our approach is orthogonal to existing state-of-the-art mitigation strategies. Thus, our method can be applied to existing works and has achieved a significant improvement.}
  \vspace{1em}
    \begin{tabular}{ll|ccc}
    \specialrule{\heavyrulewidth}{0pt}{0pt} 
    \hline
    \multicolumn{2}{l|}{Method} & Sim Score↓ & Clip Score↑ & FID↓ \bigstrut\\
    \hline
    \hline
    \multicolumn{2}{l|}{RT~\cite{somepalli2024understanding}} & 0.524  & 29.54 & 18.7  \bigstrut[t]\\
    \multicolumn{2}{l|}{RT~\cite{somepalli2024understanding} + Ours}& \textbf{0.325}  & \textbf{30.83}  & \textbf{16.7}  \bigstrut[b]\\
    \hline
    \multicolumn{2}{l|}{CWR~\cite{somepalli2024understanding}} & 0.576  & 30.13 & 18.1  \bigstrut[t]\\
    \multicolumn{2}{l|}{ CWR~\cite{somepalli2024understanding}+ Ours} & \textbf{0.343}  & \textbf{30.52}  & \textbf{16.7}  \bigstrut[b]\\
    \hline
    \multicolumn{2}{l|}{GNI~\cite{somepalli2024understanding}} & 0.615  & 30.32 & 18.9  \bigstrut[t]\\
    \multicolumn{2}{l|}{GNI~\cite{somepalli2024understanding}+ Ours} & \textbf{0.337}  & \textbf{30.86}  & \textbf{17.0}  \bigstrut[b]\\
    \hline
    \multicolumn{2}{l|}{Wen \MakeLowercase{\textit{et al.}}~\cite{wen2023detecting}} & 0.352  & \textbf{28.56} & 25.7\bigstrut[t]\\
    \multicolumn{2}{l|}{Wen \MakeLowercase{\textit{et al.}}~\cite{wen2023detecting}+ Ours} & \textbf{0.272} & 28.50  & \textbf{21.5}  \bigstrut[b]\\
    \hline
    \specialrule{\heavyrulewidth}{0pt}{0pt} 
    \end{tabular}%
  \label{tab:addinference}%
  \vspace{-1pt}
\end{table}%

\begin{table}
\centering
  \caption{Parameters impact on experimental results.}
  \vspace{1em}
  {
  \fontsize{7.5}{8}\selectfont %
    \begin{tabular}{lc|ccc|c}
    \specialrule{\heavyrulewidth}{0pt}{0pt} 
    \hline
    \multicolumn{2}{c|}{\multirow{2}[3]{*}{Parameters}} & \multicolumn{4}{c}{CIFAR-10} \bigstrut[t]\\
\cline{3-6}    \multicolumn{2}{c|}{} & MQ$_{0.4}$ & MQ$_{0.5}$ & MQ$_{0.6}$$\downarrow$ & FID$\downarrow$ \bigstrut\\
    \hline
    \hline
    \multirow{5}[0]{*}{Number of Shards $K$} 
    
    & 1    & 111 & 465 & 2030   & 8.81 \bigstrut[t]\\
    & 2     & 14    & 79   & \textbf{501} & \textbf{7.47}\\
     & 5     & \textbf{6}    & \textbf{51}   & 507 & 8.17\\
     & 10    & 10 &73 & 623   & 8.33 \\
        & 15    & 21    & 118   & 747  & 10.68 \bigstrut[b]\\
    \hline
    \multirow{3}[2]{*}{Epoches per Interaction $E$}  & 25    & 21    & 128   & 793  & 10.89 \bigstrut[t]\\
    & 50   & \textbf{10} & \textbf{73} & \textbf{623} & \textbf{8.33} \\
    & 100    & 21    & 123   & 828  & 9.57  \bigstrut[b]\\
    \hline
    \multirow{3}[2]{*}{Redistribute Proportion $P$} & 0.10   & 12    & 86   & 638  & 9.03 \bigstrut[t]\\
     & 0.25   & \textbf{10} & \textbf{73} & \textbf{623} & \textbf{8.33} \\
          & 0.50   & 11   & 90   & 750   & 8.66 \bigstrut[b]\\
    \hline
    \multirow{3}[2]{*}{Skipping Threshold ${\lambda}$}  & 0.4   & 19    & 135   & 985  & 8.69 \bigstrut[t] \\
     & 0.5     & 10    & 73   & 623   & \textbf{8.33} \\
     & 0.6   & \textbf{9} & \textbf{52} & \textbf{372} & 12.91 \bigstrut[b]\\
    \hline
    \specialrule{\heavyrulewidth}{0pt}{0pt} 
    \end{tabular}%
    }
  \label{tab:parameters_effect}%
\end{table}%
\subsection{Exploring Parameters Impact on Experimental Results.}
\label{sec:impact}
In this study, we examine how various parameters affect our experimental outcomes.
By systematically varying these parameters, we aim to understand how they influence our results and to identify the optimal settings for our experiments. 
Specifically, we conduct a series of experiments where we change the number of shards $K$, training epochs per interaction period $E$, redistribute proportion $P$, and skipping threshold $\lambda$. For each variation, we measure the impact on MQ and FID. The default parameters are set with the number of shards $K$ as 10, epochs per shard $E$ as 50, redistribute proportion $P$ as 0.25, and skipping threshold $\lambda$ as 0.5.

The results are reported in \cref{tab:parameters_effect}.

\textbf{Number of Shards $K$.} We investigate the impact of the number of data shards on model performance by setting it to 1, 2, 5, 10, and 15. $K=1$ means the default training strategy of diffusion models. Results show that the MQ scores of $K=2, 5, 10, 15$ are all lower than $K=1$, indicating that using our IET method can effectively reduce memorization. When $K=5$, the MQ score achieves the best performance. Moreover, the effect of improving image quality becomes more pronounced as the number of data shards decreases.

\textbf{Training Epochs per Interaction Period $E$.} 
We conduct experiments by varying the number of epochs per model interaction period, \MakeLowercase{\textit{i.e.}}, the interaction frequency of parameter aggregation and sample redistribute. Results show that both high and low frequencies of aggregation will reduce the performance of the memorization mitigation and image quality. Thus, we choose $E=50$ to optimize the performance of MQ.

\gxlnote{\textbf{Redistribute Proportion $P$.}
We explore the impact of the redistribute proportion parameter by setting it to 0.10, 0.25, and 0.50. As observed, when the proportion of redistributed data is too small, many frequently skipped samples cannot be relearned, limiting improvements in image quality. However, if too much data is redistributed across shards, there is a risk of exacerbating memorization. In our experimental setup, the redistribute proportion of 0.25 yields the best results.}

\textbf{Skipping Threshold $\lambda$.} We evaluate the importance of $\lambda$ in mitigating the memorization effect by setting the values of $\lambda$ to $0.4$, $0.5$, and $0.6$. A large threshold means skipping more training samples that are easily memorized. Results in ~\cref{tab:parameters_effect} show as $\lambda$ grows, more memorable training samples are skipped and the memorization phenomenon is further reduced. However, skipping more samples will reduce the model performance, \MakeLowercase{\textit{i.e.}}, the generation quality. \gxlnote{Therefore, when selecting the skip threshold, we need to strike a balance between image quality and mitigating memorization.}

\section{Conclusion}
\gxlnote{This paper presents a novel and effective training method aimed at mitigating the memorization problem in diffusion models. By analyzing the relationship between training loss and memorization, we apply different treatments to samples based on their degree of memorization, minimizing the risk of memorization. Additionally, considering that model directly learning from data can increase the likelihood of memorization and the same data may have
different interpretations on different shards, we employ several data shards to train multiple proxy diffusion models. Through multiple proxy diffusion models aggregation and redistribution of easily memorable samples cross shards, we obtain the final model, achieving a balance between mitigating memorization and maintaining image quality. We experimentally show that our method performs favorably with many existing related methods in different scenarios and datasets.}
We firmly believe that this training strategy has a broad application prospect and great development potential in the field of data privacy protection.

\bibliographystyle{IEEEtran}

\bibliography{arxiv}

\begin{IEEEbiography}[{\includegraphics[width=1in,height=1.25in,clip,keepaspectratio]{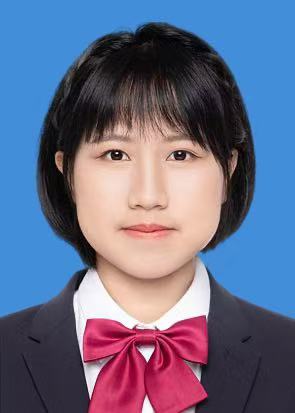}}]{Xiaoliu Guan}
received a bachelor's degree from the School of Computer Science, Wuhan University (Wuhan, China) in 2024. She is currently pursuing a master’s degree in computer science at the School of Computer Science, Wuhan University (Wuhan, China). Her research interests mainly include data privacy in diffusion models.
\end{IEEEbiography} 

\begin{IEEEbiography}[{\includegraphics[width=1in,height=1.25in,clip,keepaspectratio]{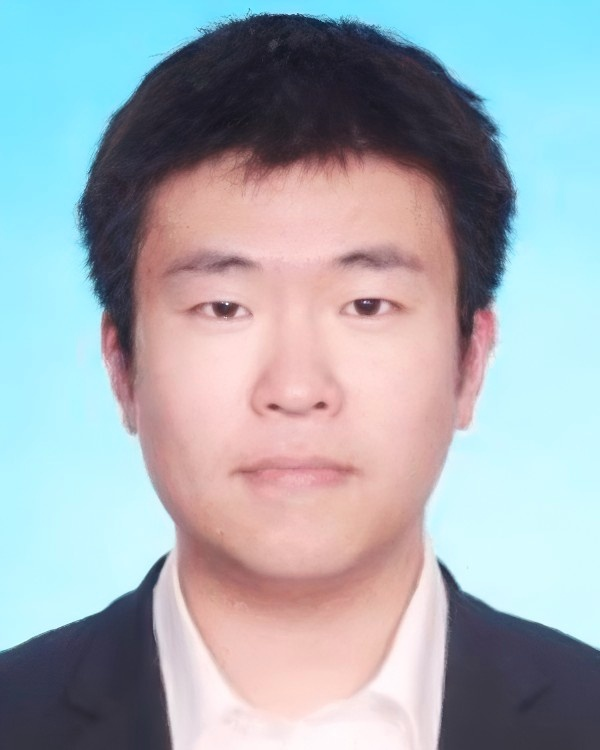}}]{Yu Wu}
 is a Professor with the School of Computer Science at Wuhan University, China. He received his Ph.D. degree from the University of Technology Sydney, Australia in 2021. From 2021 to 2022, he was a postdoc at Princeton University. His research interests are controllable generation and multi-modal perception. He was the recipient of AAAI New Faculty Highlight 2024 and Google PhD Fellowship 2020. He served as the Area Chair for CVPR, ICCV, ECCV, and NeurIPS, and also served as the Workshop Chair of CVPR 2023.
\end{IEEEbiography}
\begin{IEEEbiography}[{\includegraphics[width=1in,height=1.25in,clip,keepaspectratio]{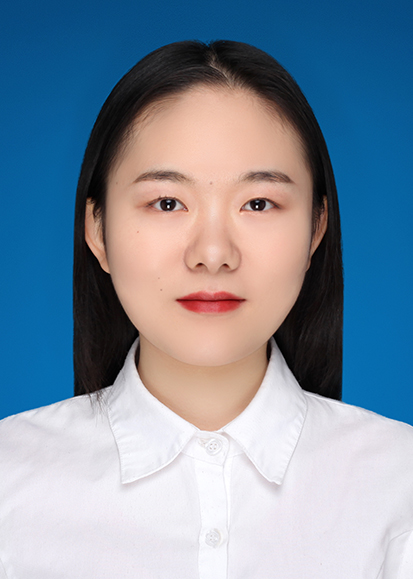}}]{Huayang Huang}
 received the master's degree from the School of Cyber Science and Engineering, Wuhan University (Wuhan, China), in 2024. She is currently working toward the PhD degree with the School of Computer Science, Wuhan University (Wuhan, China). Her research interests focus on safe generative AI.
\end{IEEEbiography}
\begin{IEEEbiography}[{\includegraphics[width=1in,height=1.25in,clip,keepaspectratio]{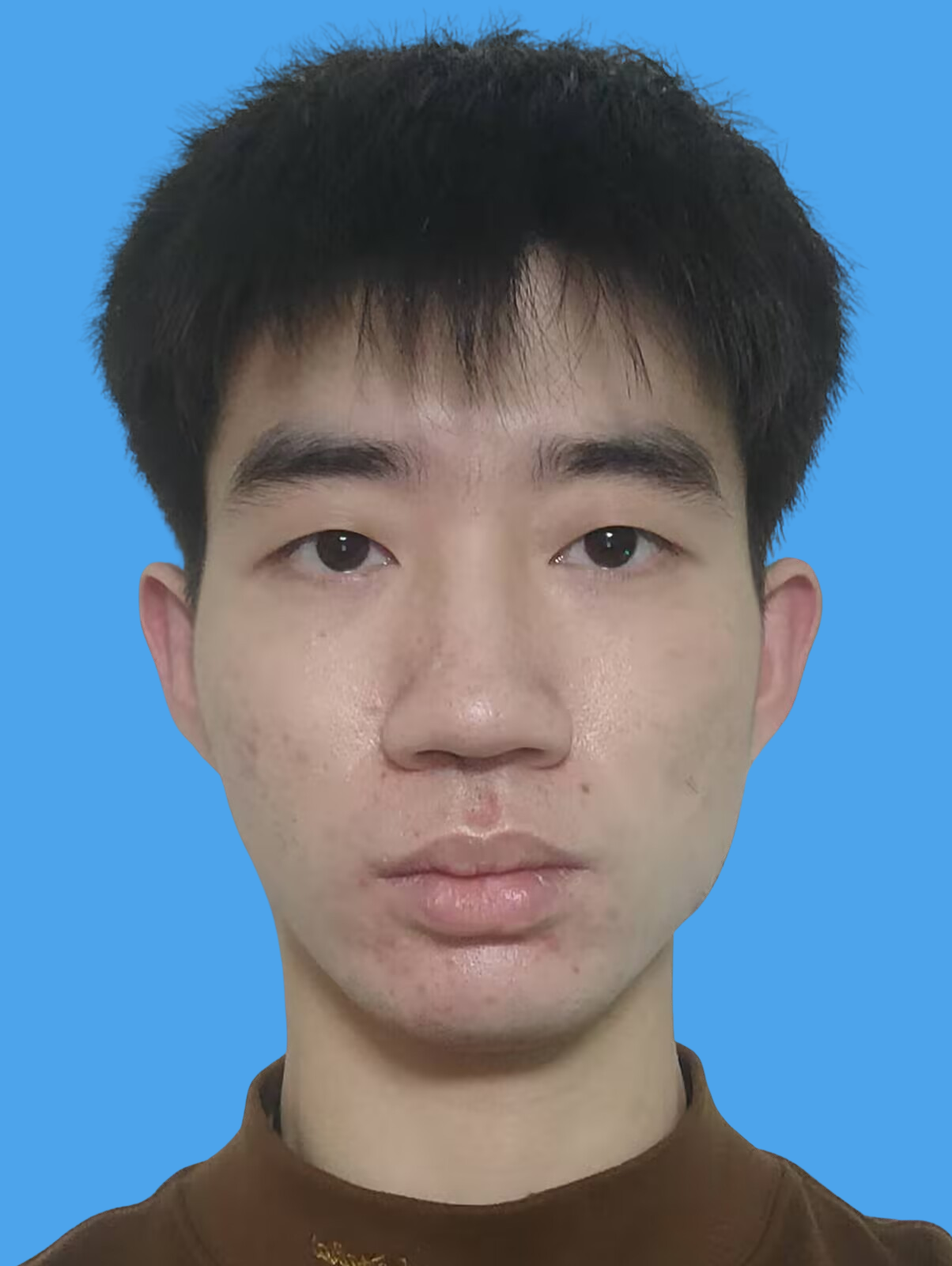}}]{Xiao Liu}
received a bachelor's degree from the School of Computer Science, Wuhan University (Wuhan, China) in 2024. He is currently pursuing a master's degree in computer science at the  School of Cyber Science and Engineering, Wuhan University (Wuhan, China).  His research interests mainly include diffusion models and multimodal learning.
\end{IEEEbiography} 
\begin{IEEEbiography}[{\includegraphics[width=1in,height=1.25in,clip,keepaspectratio]{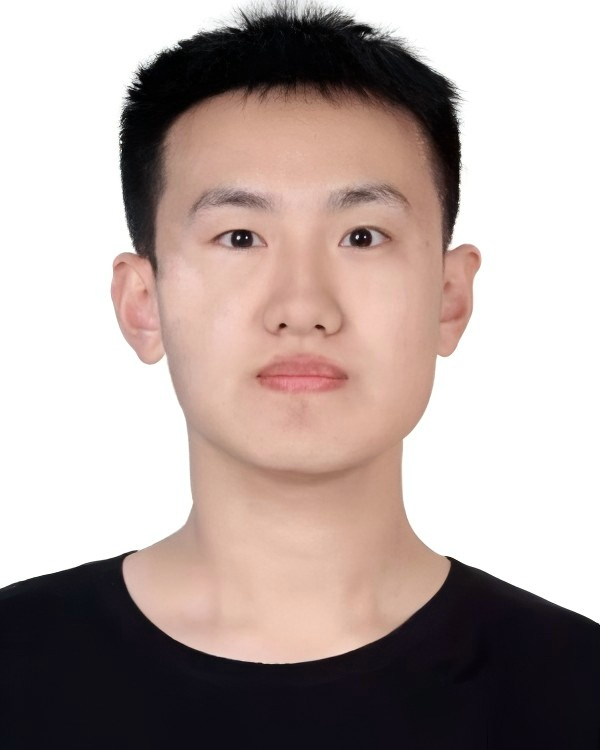}}]{Jiaxu Miao}
received the PhD degree from the University of Technology Sydney, in 2021. He is an assistant professor with the School of Cyber Science and Technology, Sun Yat-sen University, (Shenzhen, China).
His research interests include visual safety and understanding.
\end{IEEEbiography}
\begin{IEEEbiography}[{\includegraphics[width=1in,height=1.25in,clip,keepaspectratio]{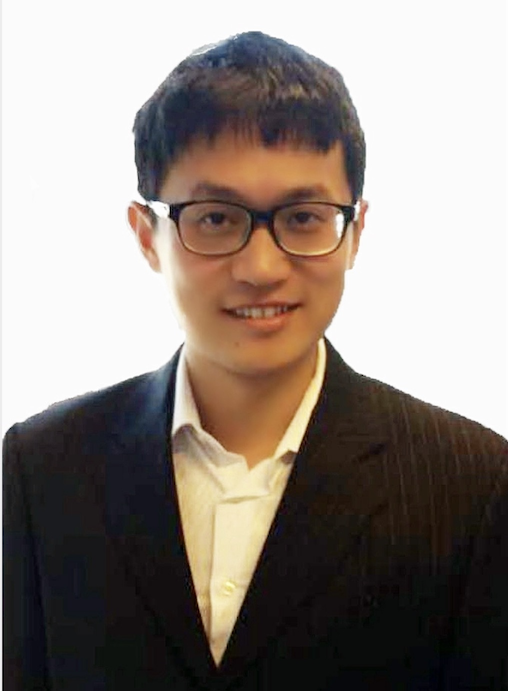}}]{Yi Yang}
(Senior Member, IEEE) is a distinguished Professor with the College of Computer Science and Technology, Zhejiang University. He has authored over 200 papers in top-tier journals and conferences. His papers have received over 70,000 citations, with an H-index of 128. He has received more than 10 international awards in the field of AI, such as the Zhejiang Provincial Science Award First Prize, the Australian Research Council Discovery Early Career Research Award, the Australian Computer Society Gold Digital Disruptor Award, and the Google Faculty Research Award. His current research interests include machine learning and its applications to multimedia content analysis and computer vision, such as multimedia retrieval and generation understanding.
\end{IEEEbiography}
\clearpage
\newpage

\end{document}